\documentclass[nohyperref]{article}

\usepackage{microtype}
\usepackage{graphicx}
\usepackage{subfigure}
\usepackage{booktabs} 

\PassOptionsToPackage{hyphens}{url}\usepackage{hyperref}

\usepackage[accepted]{icml2023}

\usepackage{amsmath}
\usepackage{amssymb}
\usepackage{mathtools}
\usepackage{amsthm}
\usepackage{svg}
\usepackage{bbm}
\usepackage{adjustbox}
\usepackage{pgf, tikz}

\usetikzlibrary{arrows,automata}
\newcommand{\defeq}{\vcentcolon=}
\newcommand{\new}[1]{\textcolor{black}{#1}}

\usepackage{color}
\usepackage{xcolor,colortbl}

\newcommand{\er}[1]{\mathbb{R}}

\usepackage[capitalize,noabbrev]{cleveref}

\theoremstyle{plain}
\newtheorem{theorem}{Theorem}[section]
\newtheorem{proposition}[theorem]{Proposition}

\theoremstyle{definition}

\theoremstyle{remark}

\usepackage[textsize=tiny]{todonotes}

\icmltitlerunning{Nonlinear Advantage: Trained Networks Might Not Be As Complex as You Think}

\begin{document}

\twocolumn[
\icmltitle{Nonlinear Advantage: Trained Networks Might Not Be As Complex as You Think}



\icmlsetsymbol{equal}{*}

\begin{icmlauthorlist}
\icmlauthor{Christian H.X. Ali Mehmeti-Göpel}{yyy}
\icmlauthor{Jan Disselhoff}{yyy}
\end{icmlauthorlist}

\icmlaffiliation{yyy}{Department of Computer Science, University of Mainz, Germany}

\icmlcorrespondingauthor{Christian H.X. Ali Mehmeti-Göpel}{chalimeh@uni-mainz.de}

\icmlkeywords{Machine Learning, ICML}

\vskip 0.3in
]



\printAffiliationsAndNotice{}  
\begin{abstract}
	We perform an empirical study of the behaviour of deep networks when fully linearizing some of its feature channels through a sparsity prior on the overall number of nonlinear units in the network. In experiments on image classification and machine translation tasks, we investigate \new{how much we can simplify the network function towards linearity before performance collapses. First, we observe} a significant performance gap when reducing nonlinearity in the network function \textit{early} on as opposed to \textit{late} in training, in-line with recent observations on the time-evolution of the data-dependent NTK. 
	\new{Second, we find that} after training, we are able to linearize a significant number of nonlinear units while maintaining a high performance, indicating that much of a network's expressivity remains unused but helps gradient descent in early stages of training. To characterize the depth of the resulting partially linearized network, we introduce a measure called average path length, representing the average number of active nonlinearities encountered along a path in the network graph. 	
	 Under sparsity pressure, we find that the remaining nonlinear units organize into distinct structures, forming core-networks of near constant effective depth and width, which in turn depend on task difficulty. 
\end{abstract}

\section{Introduction}
Deep learning as such is based on the idea that concatenations of (suitably chosen) nonlinear functions increase expressivity so that complex pattern modeling and recognition problems can be solved. While initial approaches such as AlexNet \citep{alexnet} only used moderate depth, improvements such as batch normalization \cite{batchnorm,bn_covshift,bn_autotune} or residual connections \cite{resnet} made the training of networks with hundreds or even thousands of layers possible, and this did contribute to significant practical gains.

From a theoretical perspective, a network's depth (along with its width) upper bounds the complexity of the function that it can represent \citep{vc_bounds} and therefore upper bounds the network's expressivity. Indeed, deeper networks tend to enhance performance \citep{efficientnet}, but gains seem to taper off and saturate with increasing depth. Very deep networks are also known to be more difficult to analyze  \citep{generalization}, more computationally expensive to train or infer and suffer from numerous stability issues such as vanishing \citep{vanishing_gradients}, exploding \citep{fixup} and shattering \citep{shattering_gradients} gradients. Similar arguments can be made about layer width: a sufficiently large network can memorize any given function \citep{original_uat}, but under standard initialization and training infinite width networks degrade to Gaussian processes \cite{Jacot2018}.

From a practical perspective, there are now many successful recipes for creating networks of a prescribed depth, but it is still difficult to understand -- empirically or analytically -- how many nonlinear layers and features per layer are actually needed to solve a problem, and how effective a chosen architecture actually is in exploiting its expressive potential in the sense of a deep stack of concatenated nonlinear computations.

Our paper addresses this question from an empirical perspective: we use a simple setup that associates every nonlinear unit with a cost at channel granularity, which can be raised continuously, while simultaneously trying to maintain performance. As PReLU activations \citep{prelu} can be used to interpolate continuously between a ReLU function and a linear function \citep{ringing_relus}, we replace each ReLU layer with channel-wise PReLU activations and regularize their slopes towards linearity. Such a linearized feature channel therefore only forms linear combinations of existing feature channels, thereby not effectively contributing to the nonlinear complexity of the network. To measure the nonlinear complexity or  "effective depth" of the resulting partially linearized networks, we introduce a metric called \textbf{average path length (APL)}: the average amount of nonlinear units a given input traverses until it passes the final layer. We thereby disregard subsequent linear mappings along computational paths, as these do not increase expressivity in a nonlinear sense. 

Using this tool, we make a series of experiments on common convolutional and transformer architectures on standard computer vision and machine translation tasks by partially linearizing networks with regard to the network's inputs at different stages of training and find that networks linearized later on in training obtain a significantly higher performance than networks linearized earlier on in training. This is non-trivial, since the network's expressivity is the same whether it is linearized early or late in training. We find the biggest differences in the early training phase, complementarily to the findings of \cite{nonlin_advantage} that establish a similar, but much less surprising effect when fully linearizing the network with regard to the network's weights.
	
Analyzing these partially linearized networks extracted after training, we find that we can extract very shallow networks with a surprisingly high performance for their effective depth. These findings are consistent with the lottery ticket hypothesis \citep{lottery} that there is a core nonlinear structure in networks, and with the subsequent findings of \citet{eb_lottery} that it forms within the first epochs of training. Our method allows us to compute an approximate lower bound for depth and width that the networks needs to solve a task before performance collapses and we find that these are approximately constant for a given task and regularization strength, independently of the width and depth of the initial network. We also find that the effective depth of this core nonlinear structure grows with problem complexity for a fixed regularization strength.

\section{Related Work and Contributions}
Different approaches to network pruning were explored in recent years: magnitude-based weight pruning, weight-regularization techniques, sensitivity-based pruning and search-based approaches \citep{compression_survey}. \citet{lottery} extract a highly performant, sparse and re-trainable subnetwork by removing all low-magnitude weights after a given training time, re-initializing the network and iterating this process. This motivates the "lottery ticket hypothesis" of a network consisting of a smaller core structure embedded in the larger, overparametrized and redundant network, which, in their case, can be extracted by weight pruning. Our paper prunes nonlinear units instead of weights, but comes to similar findings of a problem-difficulty-dependent minimal set, embedded in a much larger and deeper network, when considering nested nonlinear computations. \citet{eb_lottery} claim that the final accuracy of lottery tickets drawn after at early training is already drastically higher than at initialization. \citet{sanity_lottery} conduct "sanity checks" on the lottery ticket hypothesis and conclude that only the number of remaining weights matters for a given dataset. We conduct similar checks and find that transferring simple statistics such as how many nonlinear units are active per layer are not sufficient to recover full performance.

Simplification of networks by reducing nonlinearity has become a major area of interest. A lot of recent work has studied the neural tangent kernel (NTK) approximation, which linearizes the network function with regard to its parameters.  It arises in the infinite width limit (under mild conditions) or by explicitly performing a linear Taylor-approximation of a finite network~\cite{Jacot2018,nonlin_advantage}. As it fully linearizes training, the NTK has been tremendously useful for gaining a better understanding of the training of deep network, such explaining double-descent generalization \cite{belkin2019reconciling,wilson2020bayesian}. Maybe unsurprisingly, linearized training hurts performance in practice~\cite{nonlin_advantage} and theory: \citet{Roberts2022} attribute it to the loss of detection of higher-order moments in the data distribution). Fort et al.~have coined the term \textit{nonlinear advantage} for the observed discrepancy in performance between the network's NTK and the nonlinear network function that vanishes over time when training with low learning rate. Within the NTK framework, the impact of ReLUs can be captured by path kernels \cite{Lakshminarayanan2020}, the learning of which improves results and generalizes when retraining, and can be used to understand pruning methods at initialization \cite{pruning_init}. Our APL measures are tightly related to the proposed (gated) path-integral formulation there. Our paper simplifies the network function itself by reducing the number of nonlinear units in the network and therefore partially linearizing it in both inputs and weights, finding a similar, but difficulty-dependent nonlinear advantage.

\citet{layer_folding} use a methodology similar to ours, but applied layer-wise and aiming at improved performance characteristics at inference time. Our channel-wise approach allows us to reduce significantly more nonlinear units in the network whilst maintaining a similar performance as well as characterize the "effective width" of the emerging core network. A \new{training time} dependent effect as we show it, is not studied by the authors.



\section{Reducing Nonlinear Feature Channels}
\label{sec:regularizer}
In order to reduce the amount of nonlinear feature channels in a network, we take network architectures and replace their ReLU activations with PReLUs. We then use a single PReLU weight for every channel and add a sparsity regularization of $L_{0.5}= \sum |1 - \alpha_i|^{0.5}$ to the regular training loss scaled with a \textit{regularization weight} $\omega$, where $\alpha_i$ is the variable slope of the i-th PReLU. We chose channel-wise PReLU units because the latter allows a much bigger reduction in overall nonlinearity compared to layer-wise units, and pixel-wise PReLUs would entail an unreasonable amount of additional parameters. Since the regularization loss term is discontinuous at $1$, we disable a PReLU unit if their slope gets close enough to one. We call such a unit \textit{inactive}, while all other units are \textit{active}.

By regularizing PReLU units this way, the slope of inactive units is locked to $\alpha_i=1$, but the slope of active units is an arbitrary number between 0 and 1. After reaching a goal percentage of disabled PReLU units, it is possible to regularize the slope of the remaining active units back to 0, effectively transforming the network back to a regular ReLU network and relating to \citep{few_patterns}, but we found that this method can be hurtful to performance and therefore refrain from using it.

\subsection{Average Path Length (APL)}
\begin{figure}[h!]
	\centering
	\includegraphics[width=.7\linewidth]{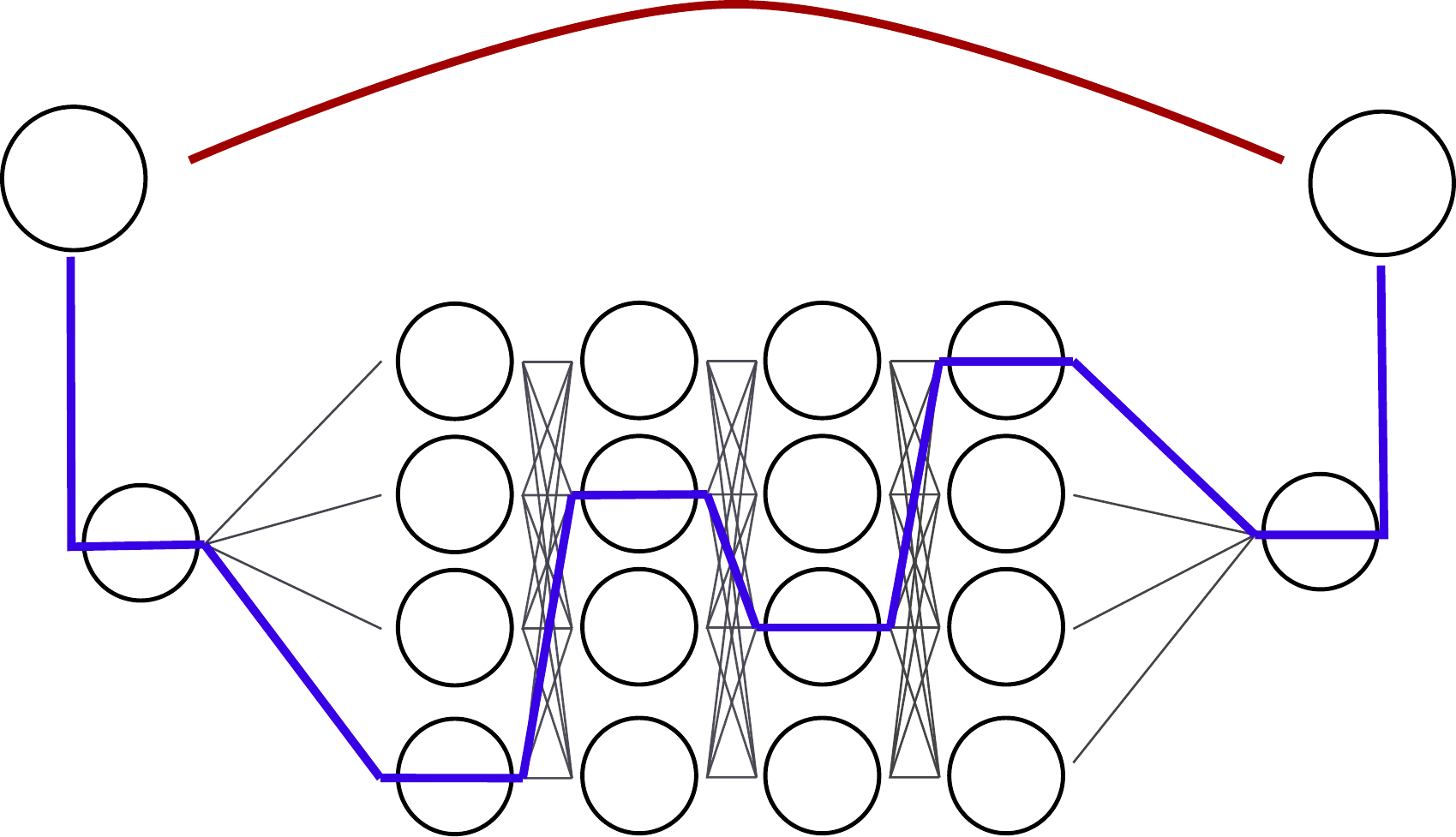}
	
	\caption{In this Figure, a residual connection (red) skips 4 small fully connected layers. The left-most and right-most nodes are connected by 257 different paths. Using \textit{unnormalized} average path length implies a $1/257$ chance of selecting the red path and the same chance of selecting the blue path. Using \textit{normalized} average path length, we have a $1/2$ chance of selecting the red path and a $1/512$ chance of selecting the blue path.}
	\label{apl_vs_napl}
\end{figure}

The depth of a network, according to its traditional notion, corresponds to the \textit{maximum} amount of nonlinear units encountered  when following the computation graph of a network from input to output. After partially linearizing such a network, this value remains the same, assuming at least one channel per layer remains active (non-linear).  Therefore, computing the \textit{average} amount of nonlinear units instead of the maximum seems like a more sensible characterization for partially linearized networks.

Let $G=(V,E)$ be the directed acyclic graph that represents the computation graph of a given feedforward neural network. Since we are only interested in the nonlinear structure of the graph, a node in the graph corresponds to a PReLU unit in the network. We denominate $V_d\subset V$ the subset of nodes that correspond to the $d$-th layer in the network. Let $v_{in}, v_{out} \in V$ be the respective input and output vertices of the graph. Let $R\subset V$ be a subset of the vertices that represent the blocks containing an active PReLU i.e. $|\alpha - 1| > \epsilon$, where $\alpha$ is the weight of the PReLU.

Let $\mathcal{P}^{(n)}\subset V^n$ be the set of all paths of length $n$ in $G$ that originate in $v_{in}$, i.e. $p_1 \in v_{in}$ and  $(v_i,v_{i+1})\in E$ for all $1\leq i\leq n-1$.   We define the \textit{effective path length}  of a path $p\in\mathcal{P}^{(n)}$ as the number of active PReLU activations it traverses: 
\begin{equation*}
\phi(p) \defeq |\{v_i \in p~|~v_i \in R,~  1\leq i \leq n \}|,
\end{equation*}
which is always smaller or equal to its regular length $|p|$. Let $v\in V$ be a vertex of the graph, we then define its path histogram function as the number of paths from $v_{in}$ to $v$ of effective length $l$ :
\begin{equation*}
\phi^{(l)}(v) \defeq |\{p \in \mathcal{P}^{(l')} ~|~  p_l = v, ~\phi(p) = l, ~1\leq l' \leq l \}|.
\end{equation*}
 We finally define the \textit{average path length (APL)} of the network as:
\begin{equation*}
	APL(G) \defeq\frac{\sum_{i=0}^{d}i\cdot \phi^{(i)}(v_{out})}{\sum_{i=0}^{d}\phi^{(i)}(v_{out})},
\end{equation*}
where $d$ is the depth of the network. In order to effectively compute the APL of a network, we resort to dynamic programming. 

\begin{proposition}
	Let $v\in V$ be a vertex in the network. We can then compute its path histogram function by summing over all vertices that have an outgoing edge to $v$:
	
	$\phi^{(l)}(v) = \begin{cases}
		\sum_{v'\in V}\mathbbm{1}_{(v', v)\in E}\cdot\phi^{(l-1)}(v')\quad\text{if }v\in R,\\
		\sum_{v'\in V}\mathbbm{1}_{(v', v)\in E}\cdot\phi^{(l)}(v')\quad~~~ \text{otherwise.}
	\end{cases} $ 
\end{proposition}

\begin{proof} Assume that we have the full histogram of all vertices of layer $l-1$ and lower and want to calculate the histogram of a given vertex $ v\in V$ in layer $l$. Let $V'\defeq \{v'\in V | (v', v)\in E\}$  be the set of all vertices that have an edge to $ v$. Since $G$ is a DAG, all paths from $v_{in}$ to $ v$ must go through exactly one node of $V'$. The histogram of $ v$ can therefore be decomposed as shown above, shifting if $ v$ contains an active PReLU.
\end{proof}

\textbf{Implementation details:} We implement this recursion in a modified forward pass through the network by re-using the batch dimension of the input tensor as "histogram dimension" that saves the path histogram function $\phi^{(l)}(v)$ for a given neuron $v$. By setting all weights of a linear (fully connected or convolutional) layer to one and biases to zero, executing the layer then automatically outputs for each neuron the sum of all inputs and therefore sums the path histogram functions of all incoming nodes. We then just need to "shift" the obtained histogram if an active PReLU is present to obtain the correct histogram function for $v$. Other layers such as batch normalization or pooling layers are ignored. We finally use a constant $(1,0,\dots,0)$ input for network and extract the obtained histograms in the last layer. For reasons discussed below, it can be useful to normalize the histograms before adding them inside a ResBlock; we call the resulting value  the \textit{normalized average path length}. An illustrative example for a histogram computation of a non-residual and a residual network is shown in the Appendix in Figure \ref{hist_example}.

By Proposition 1, the \textit{unnormalized average path length} (APL) describes the expected number of active PReLU units a path contains if we  draw a path \textit{uniformly from the set of all possible paths}. In networks with residual connections, this heavily favors longer paths, as every additional layer used increases the number of possible paths exponentially (ref. Appendix Figure \ref{hist_explosion}). In this measure, despite residual connections, the initial path length of a ResNet is only slightly lower than its depth which might seem unintuitive.

The \textit{normalized average path length} (NAPL) describes, as illustrated in Figure \ref{apl_vs_napl}, the expected number of active PReLUs a path contains if we follow a random outgoing edge at every node in the path. In a residual network this means that inside a ResBlock, both summands (main branch and residual connection) have equal weight.

Both APL and NAPL do not depend on the absolute number of active PReLUs in a layer but rather on their relative proportion. For this reason, we will also use the simple measure of \textit{effective network width} or ENW of a network. It is the absolute number of active PReLUs per layer, averaged over all layers. This measure depends only on the extracted "core" network and is therefore useful for comparing architectures of different width.

\section{Experiments}
In this section, we apply linearization to network architectures at different stages of training to show the existence of a discrepancy in performance between networks partially linearized earlier and later in training that we call \textbf{nonlinear advantage}. Once that we established this effect, we observe the performance and shape of the networks resulting when linearizing after training to convergence.

As our techniques requires networks with ReLU activations, we chose a ResNet \citep{resnetv2}, PyramidNet \citep{pyramidnet} with ("Short") and without ("NoShort") residual connections as well as Transformer \cite{transformer} as examples of standard architectures. For computer vision tasks, we work with standard image classification datasets of variable but well-known difficulty: CIFAR-10, CIFAR-100  \citep{cifar}, CINIC-10 \cite{cinic10}, Tiny ImageNet \citep{tiny_imagenet} and ILSVRC 2012 (called ImageNet in the following) \citep{imagenet}. As for NLP tasks, we use the Multi30k \cite{multi30k} machine translation task (german to english). We train all networks from scratch except on ImageNet where we use a pre-trained ResNet50 from the Torchvision library for linearization. 

\begin{figure*}[h!]
	\begin{minipage}[t]{0.475\linewidth}\vspace{0mm}%
		\vspace{1.35cm}%
			\begin{adjustbox}{width=\columnwidth,center}
				\begin{tabular}{lllll l}
					\toprule
					Architecture& Base & Linear.  & Exact & Layerwise P. & Global P.\\
					\midrule
					\textit{CIFAR-10}&&&&&\\
					\\
					ResNet56S&92.7 & 89.2&\cellcolor{cyan!25}$89.0 \pm 0.0013$&\cellcolor{cyan!25}$90.0 \pm 0.0021$&\cellcolor{cyan!25}$89.7 \pm 0.0027 $\\
					
					ResNet56NS&84.8 & 81.3&\cellcolor{cyan!25}$81.8 \pm 0.0033$&\cellcolor{cyan!25}$81.8 \pm 0.0065$&  \cellcolor{red!25}$74.0 \pm 0.0063$\\
					
					PyramNet110S &94.7&91.5&\cellcolor{cyan!25}$91.4 \pm 0.0003$&\cellcolor{cyan!25}$ 91.7 \pm 0.0012$&\cellcolor{cyan!25}$91.6 \pm 0.0014$\\
					\midrule
					\textit{CIFAR-100}&&&&&\\
					\\
					ResNet56S & 69.9 &68.3&\cellcolor{red!25}$66.8 \pm 0.0042$&\cellcolor{red!25}$66.7 \pm 0.0134$&\cellcolor{red!25}$65.2 \pm 0.0044$\\
					ResNet56NS & 56.7 & 54.8& \cellcolor{red!25}$45.3 \pm 0.0110$ &\cellcolor{red!25}$45.8 \pm 0.0263$& \cellcolor{red!25}$47.0 \pm 0.0218$ \\
					
					\bottomrule
				\end{tabular}
		\end{adjustbox}
		\caption{Test accuracy of the extracted partially linearized network compared to baseline networks where we re-train the same network from scratch with the same amount of inactive PReLUs but differently distributed. Standard deviation over five runs is indicated for the re-trained networks.}
		\label{retrain_table}
	\end{minipage}
	\hfill
	\begin{minipage}[t]{0.475\linewidth}\vspace{0mm}%
		\centering
		\includegraphics[width=7cm]{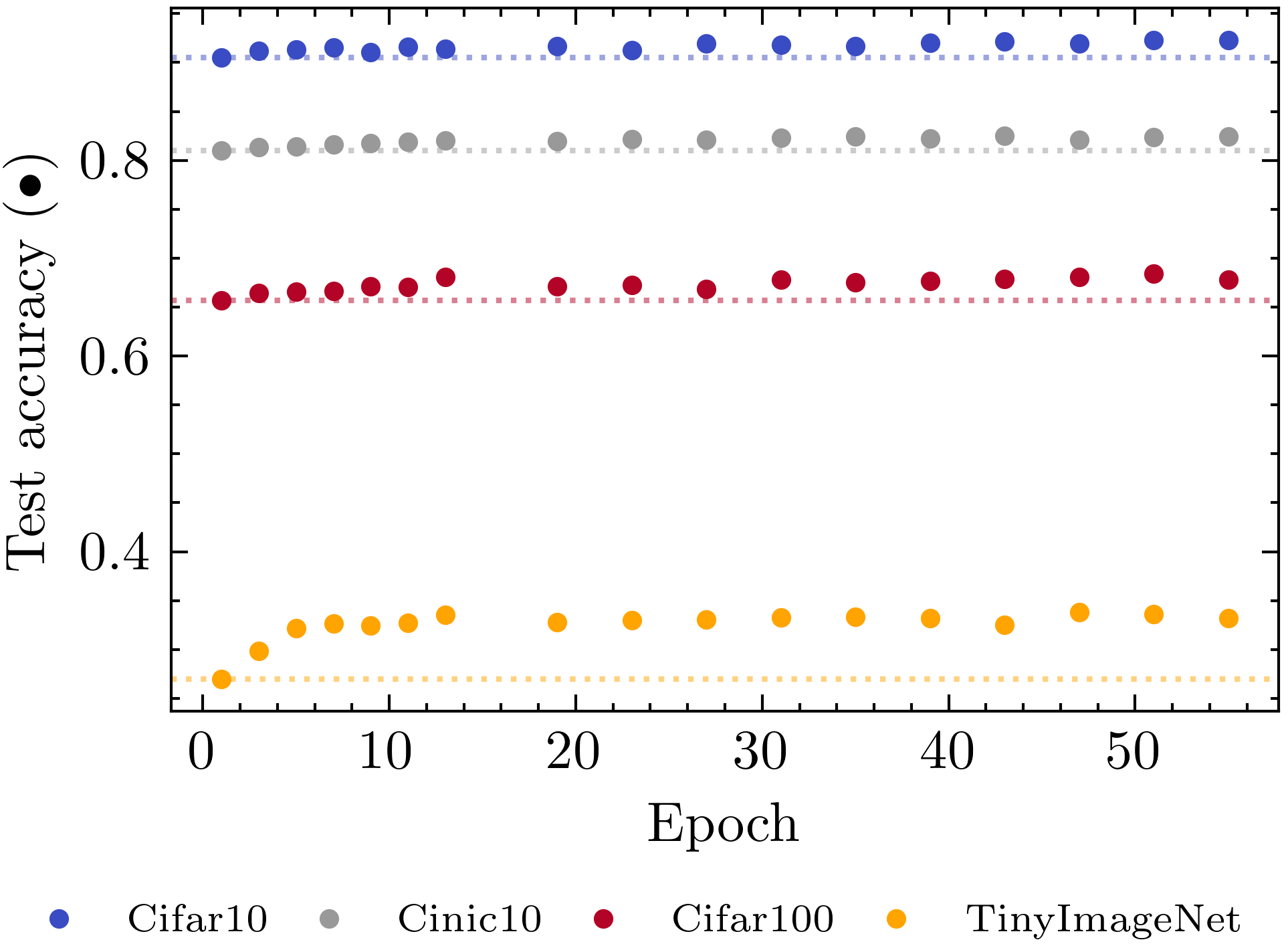}
		\caption{Test accuracy of networks partially linearized at different epochs during training on different datasets. The dotted lines indicate the height of the first data point for visual reference.}
		\label{nonlin_advantage}
	\end{minipage}
\end{figure*}

In the experiments of Figure \ref{nonlin_advantage} and \ref{nonlin_advantage_nlp}, we switch on our linearizing regularizer described in Section \ref{sec:regularizer} at the indicated time during training, in order to capture a time-dependent effect. In all other experiments, the partial linearization happens in a separate phase, after conventionally training the network:

\textbf{Training Phase:} We conventionally train the network with the most basic setup that is capable of delivering benchmark results for the chosen architectures: ReLU units, momentum SGD, a multistep learning-rate scheduler and weight decay. 

\textbf{Linearization Phase:} The ReLU units are replaced by regularized PReLUs (with initial negative slope 0) and we resume training in a shorter post-training step. PReLUs are considered inactive and frozen if their slope is higher than 0.99 (1\% margin). Concerning learning-rate scheduling in the post-training step, we need a big learning rate initially to reach the target nonlinearity and a lower learning rate afterwards in order to reach a good performance. We therefore revert to the initial learning rate and  use the same multistep scheduling as in the regular training phase adapted to the shorter post-training phase. 

Further details about architectures and training regimes used can be found in the Appendix at Section \ref{appendix:sec:arch_details}. We decided to use the normalized average path to avoid overflows for deeper networks (the absolute number of paths through the network grows exponentially in depth) and because it is in-line with previous works discussing path lengths in ResNets \citep{resnet_ensemble}. Results with unnormalized path length yield similar results albeit the absolute numbers are higher as shown in the Appendix at Section \ref{appendix:sec:further_exp}.

\subsection{The Nonlinear Advantage}
\label{sect_nonlin_advantage}
In this section, we want to establish the existence of a \textbf{nonlinear advantage} by we comparing the final performance of a network that is linearized at different stages of training. We carefully choose our experimental setup such that the difference in performance can be purely attributed to the difference in nonlinear units and not to other factors such as training time, learning rates or architectural differences.

\subsubsection{Re-training Partially Linearized Networks From Scratch}
In a first step, we consider the most extreme case of comparing a network partially linearized \textit{after being fully trained} to a network of the same architecture that contains the same amount of nonlinear units trained \textit{from scratch}. \new{We want to see whether we can train a network containing the same amount of nonlinear units as the extracted network to the performance of the latter and whether transferring simple statistics (eg. the amount of active PReLUs per layer) is sufficient to do so.}

The experimental setup is the following: we regularly train a ReLU network and linearize it in a post-training phase, replacing ReLUs by regularized PReLUs as described above. We then transfer the amount of nonlinear units to a newly initialized network and re-train the network 5 times, using the same number of epochs and schedule as in the original training phase. We also use (non-regularized) PReLU units in the network for re-training, so that the amount of parameters is comparable. Apart from the mask of inactive PReLUs, everything  else (network weights, optimizer etc.) is re-initialized (with a random seed) and the network is trained from scratch. We consider three different ways of transferring the distribution of inactive PReLU units from the partially linearized network to the new network that work at different granularity: exact, layer-wise and network-wise:

\begin{itemize}
	\itemsep 0em 
	\item \textbf{Exact: } The exact  binary masks of inactive PReLU are kept.
	\item \textbf{Layer-Wise Permutation:} The binary masks of inactive PReLUs are kept but shuffled with all PReLU units within the same layer.
	\item \textbf{Global Permutation:}  The binary masks of inactive PReLUs are kept but shuffled with all PReLU units in the network.
\end{itemize}

\begin{figure*}[t!]
	\begin{minipage}[t]{0.475\linewidth}
		\centering
		\includegraphics[width=0.9\linewidth]{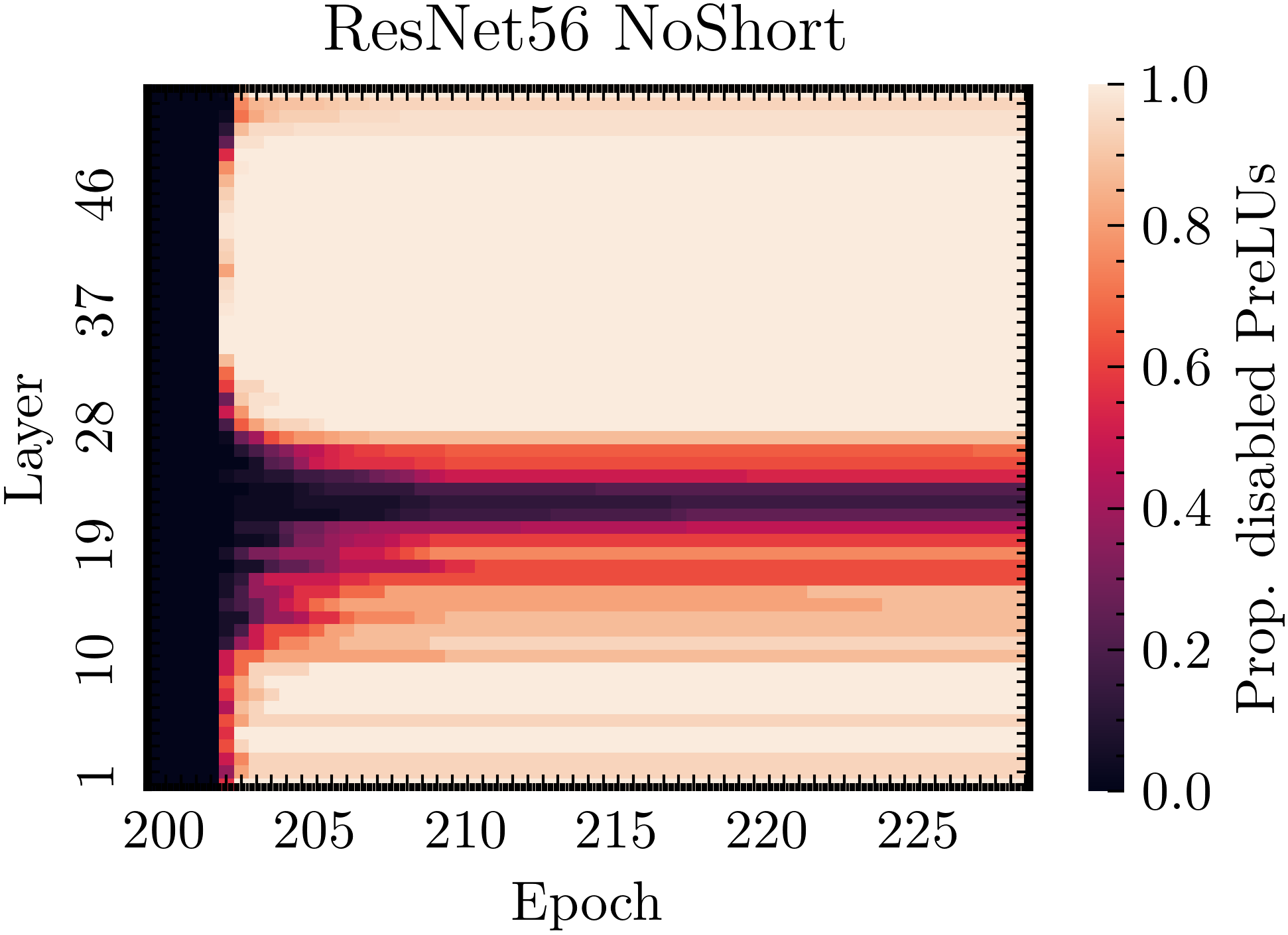}
	\end{minipage}
	\hfill
	\begin{minipage}[t]{0.475\linewidth}
		\centering
		\includegraphics[width=0.9\linewidth]{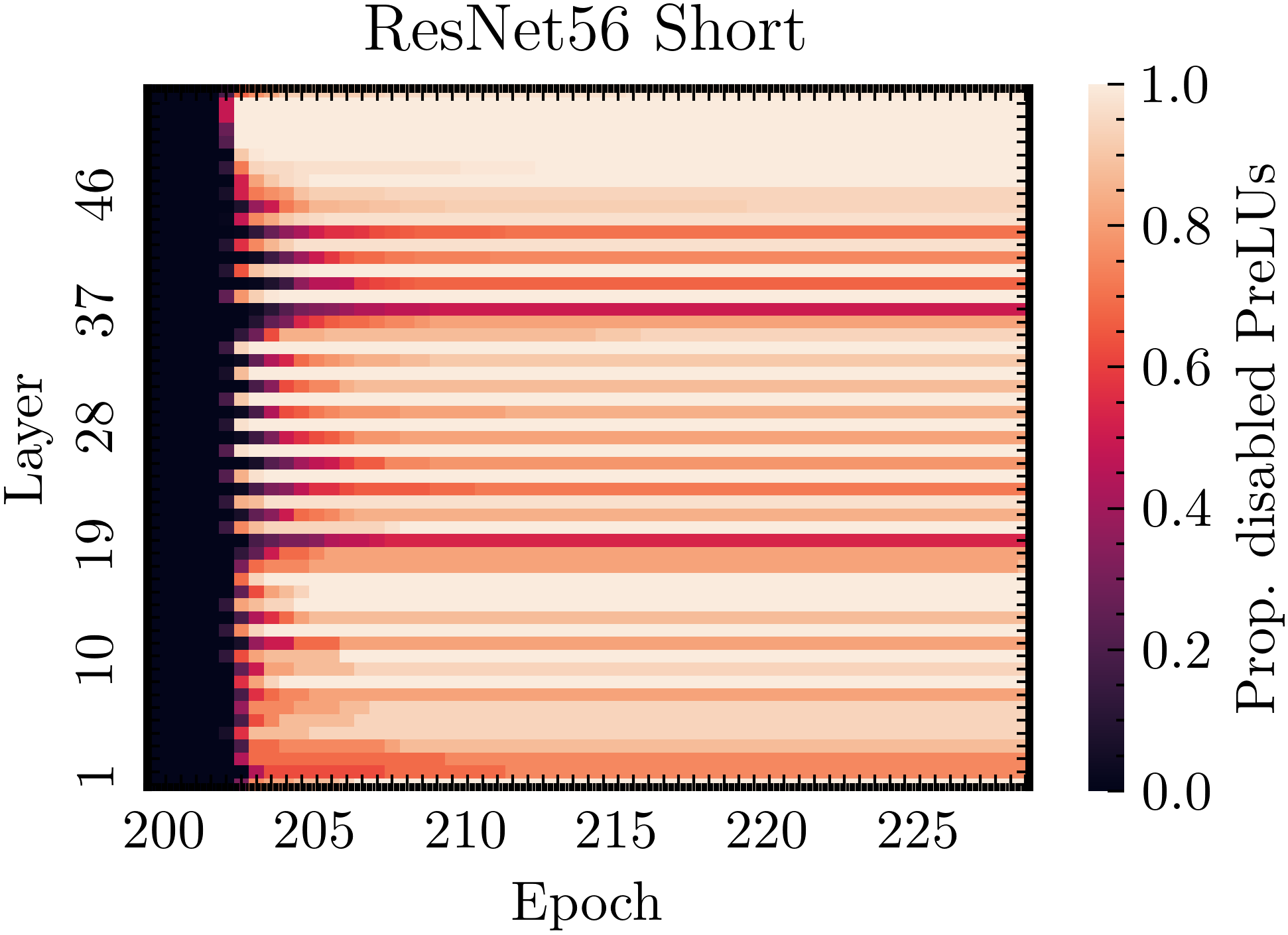}
		
	\end{minipage}
	\caption{Proportion of inactive PReLUs when partially linearizing a ResNet56 NoShort / Short with $\omega = 0.003$.}
	\label{prop_disabled}
\end{figure*}

We summarized the results in Figure \ref{retrain_table}, where we abbreviated "S" for Short and "NS" for NoShort networks. We see that for the "easy" dataset CIFAR-10, the nonlinear advantage is nonexistent since all networks trained with the exact and layerwise permutated nonlinearities reach the full performance of the network that was partly linearized after training. The slight gain in performance can be attributed to the higher number of epochs where the network can adapt to the missing nonlinearity. Further we see that only for the ResNet56 NoShort, the network which differs most from a uniform distribution in its remaining PReLU units (ref. Figure \ref{prop_disabled}), the full performance was not reached with a global permutation in PReLU masks whereas for all other architectures, full performance was reached. We conclude that nonlinear advantage is nonexistent for this easy dataset and the layerwise distribution of PReLU units matters only for networks with a very distinct (non-uniform) structure in its remaining PReLU units.
As for the significantly more difficult dataset CIFAR-100, we see that no setting can reach the performance of the network that was partly linearized after training, not even re-training where the exact PReLU masks are transferred; this \new{indicates} the existence of a nonlinear advantage for harder problems.
\subsubsection{Nonlinear advantage is stronger in early training}
In a second step, we want to break down how linearizing a network at different stages of training affects its performance. For this, we train a ResNet56 Short on datasets of varying difficulty since previous results indicate that we can only measure it on harder datasets. At different stages of the training, we activate our regularizer with a fixed regularization weight, resume training and measure the final performance of the network after a given number of epochs. As regularizing the network at different stages of training with the same regularization weight can result in massive differences in the amount of inactive PReLUs, we slightly modified our regularizer to stop when a goal percentage ($80\pm1\%$, an amount high enough to impact performance) of inactive PReLUs over all layers is reached. In order not to overshoot our goal percentage, we lower the regularization weight when close to our target. We carefully tune the regularization weight in order to avoid undershooting the target percentage. We see in Figure \ref{nonlin_advantage} that networks regularized later in training are significantly more performant than networks partially linearized earlier in training. The biggest differences are visible in the first 15 epochs of training and the effect is particularly pronounced for the harder datasets, indicating a correlation between effect strength and the hardness of the task at hand. In the Appendix in Figure \ref{nonlin_advantage_nlp} we have shown that for a transformer architecture training on a machine translation task, \new{the test perplexity of networks is higher for networks partially linearized early as opposed to later epochs. The biggest difference occurs within the first 20 epochs of training.}

\subsection{Performance and Structure of Partially Linearized Networks}

\new{Having established in the previous sections that the performance of a network linearized after training cannot simply be recovered by training with a similar amount of nonlinear units from scratch, we now want to understand better \textit{how shallow} we can make a trained network until performance collapses and \textit{where the remaining nonlinearities are located} in the network.}

\begin{figure*}[tb]
	\begin{minipage}[t]{0.475\linewidth}\vspace{0mm}%
		\centering
		\includegraphics[width=7cm]{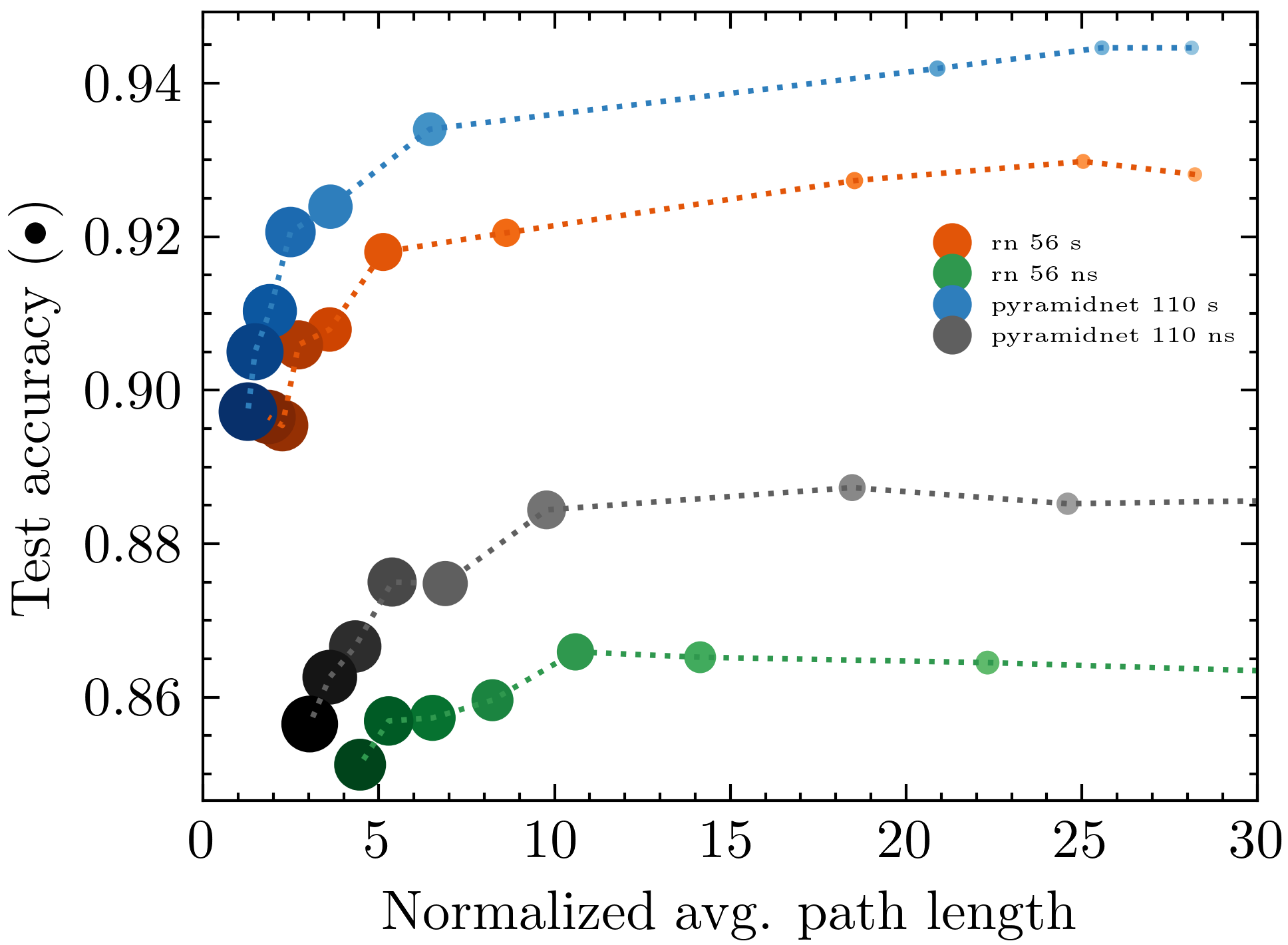}
		\caption{NAPL and test accuracy for different partially linearized network architectures with  $\omega\in [0.0005, 0.005]$.}
		\label{performance}
	\end{minipage}
	\hfill
	\begin{minipage}[t]{0.475\linewidth}\vspace{0mm}%
		\centering
		\includegraphics[width=7cm]{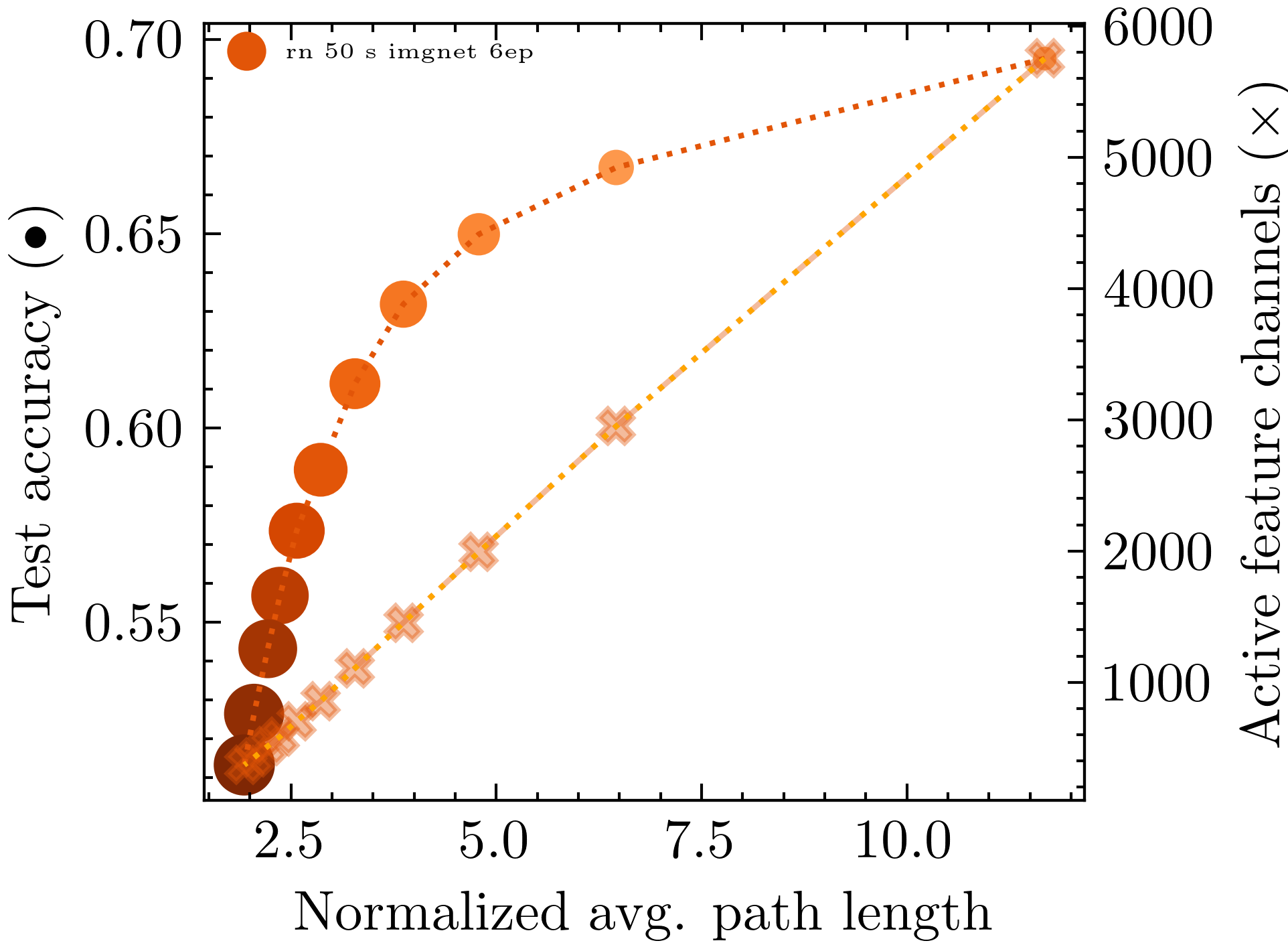}
		\caption{NAPL, accuracy and active features channels for a partially linearized ResNet50 pre-trained on Imagenet with $\omega\in [0.0005, 0.0025]$.}
		\label{performance_imgnet}
	\end{minipage}
\end{figure*}

\label{sect_performance}
We observe the temporal evolution of the linearization process for two different network architectures in Figure \ref{prop_disabled}. When evaluating the proportion of inactive PReLUs per layer, we note that every architecture presents a distinct pattern: for the ResNet56 NoShort, we see that the remaining nonlinearity is concentrated in a connected block, whereas for the ResNet56 Short, the remaining active PReLUs are distributed more evenly over the layers. Interestingly, the connected block of remaining nonlinearity in the ResNet56 NoShort is located in the middle of the network and not on either end, excluding simple vanishing/exploding gradients at initialization effects as a cause. The fact that for the ResNet56 NoShort, many layers are fully linearized \textit{without explicit incentive to do so} \new{might indicate} that such network architectures might not use their full expressive potential. The stripe-like structure of remaining nonlinearities in the ResNet56 Short corresponds to the placement of the residual connections and indicates that these might help in utilizing the full depth of the network. We found the qualitative behavior for both architectures to be consistent on the CIFAR-100 dataset (ref. Appendix), albeit the exact location of the connected nonlinear layer block changes. We conclude that despite regularizing every nonlinear unit equally, distinct patterns form in the remaining nonlinearities in the network that depend on network architecture.

Second, we want to demonstrate the effects of partial linearization on generalization performance for different architectures and datasets. We plotted the performance of partially linearized networks  for different choices of $\omega$ on Cifar10 in Figure \ref{performance} (ref. Appendix Figure \ref{regtest_archcompare_cifar100} for Cifar100). Darker colors represent a higher regularization weight and the disk sizes represents the global proportion of inactive PReLUs. We can see that for all networks, the top-1 test performance remains high even for a comparably small NAPL values until it collapses. We also see that depending on network architecture, for a similar NAPL value, different networks architectures present a distinct percentage of inactive PReLUs, further supporting our claim of a network-dependent structure being extracted by linearization. A similar plot showing explicitly the proportion of active PReLU units instead of NAPL can be found in the Appendix in Figure \ref{regtest_archcompare_active} (resp. \ref{regtest_archcompare_active_cifar100} for Cifar100) and shows qualitatively the same behavior. We further verified our claims for a ResNet50 on the ImageNet dataset in Figure \ref{performance_imgnet}; note that this network contains a non-ReLU activation layer (maximum pooling) that we included in our calculations by increasing all to the NAPL values by one.


\subsection{Analyzing the Shape of the "Core Network"}

In this section, we investigate whether we can find some regularities in the shape of the resulting network if we partially linearize networks of different initial shape.
\begin{figure*}[tb]
	\begin{minipage}[t]{0.475\linewidth}\vspace{0mm}%
		\centering
		\includegraphics[width=7cm]{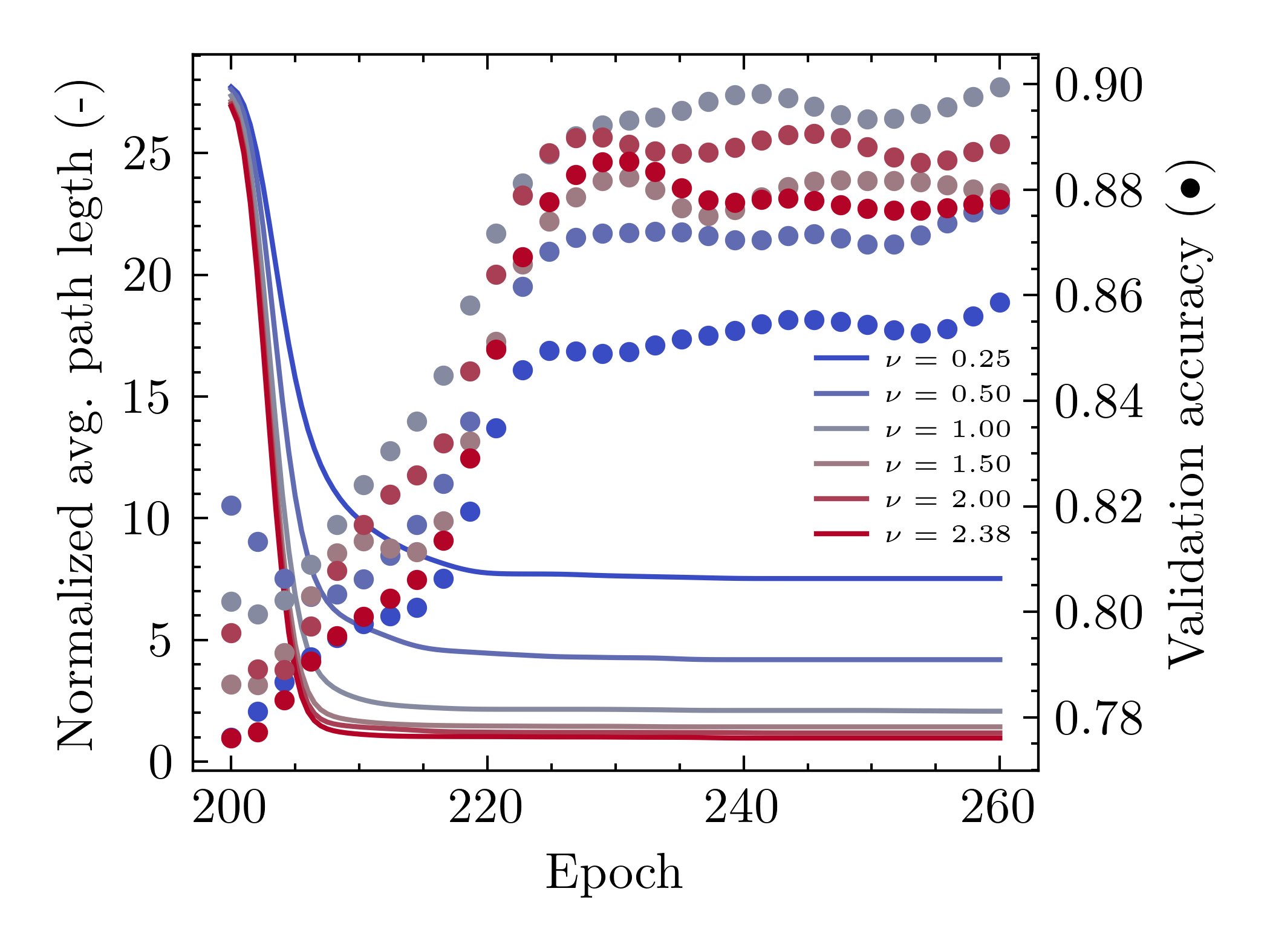}
		\caption{Validation accuracy and average path length during linearization of ResNets56 of different width for $\omega = 0.003$.}
		\label{widthtest}
	\end{minipage}
	\hfill
		\begin{minipage}[t]{0.475\linewidth}\vspace{0mm}%
		\centering
		\includegraphics[width=7cm]{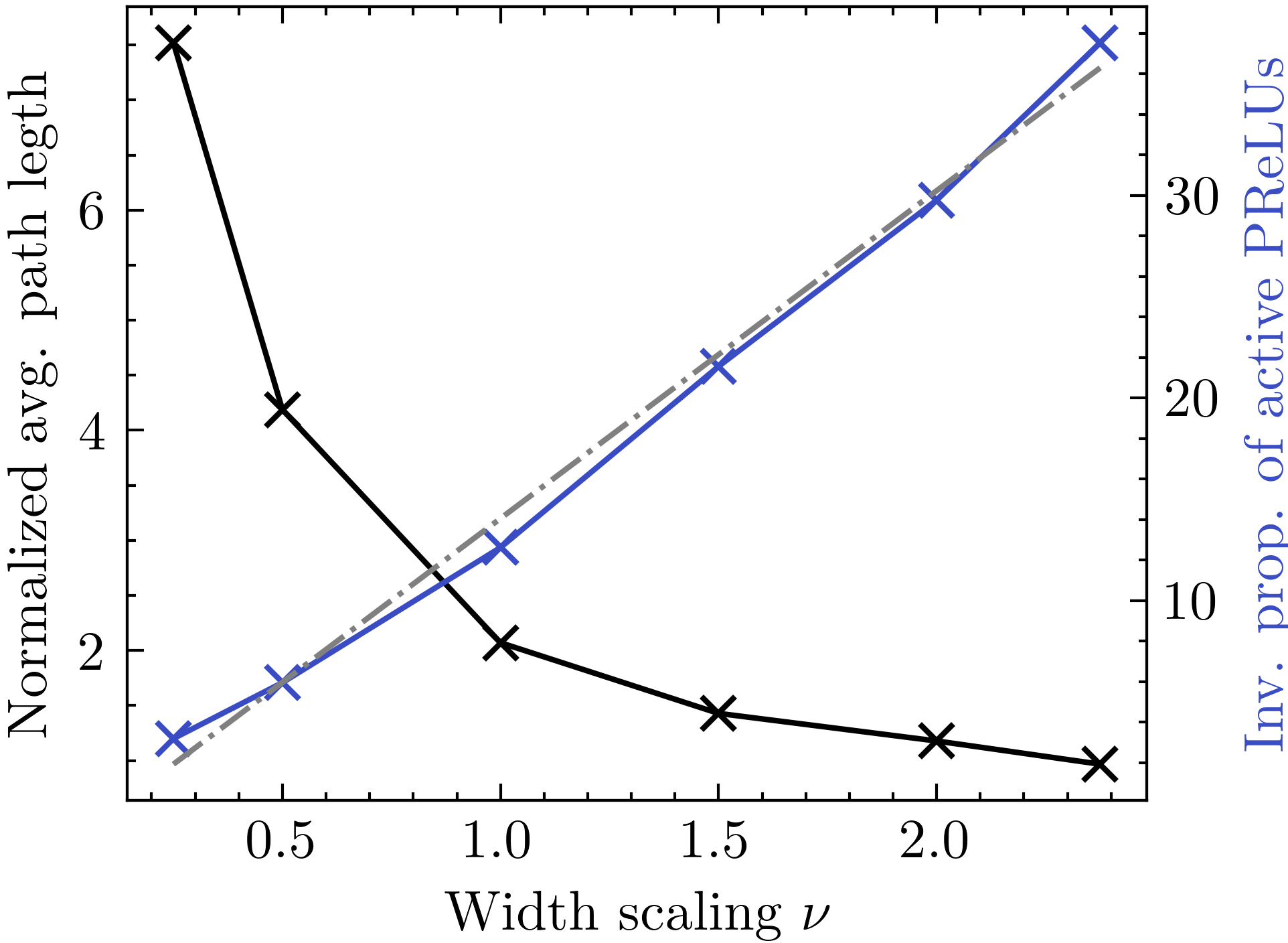}
		\caption{Inverse proportion of active PReLUs and NAPL for ResNet56 Short of different width after linearization for $\omega = 0.003$. A linear fit to the blue curve is drawn in grey.}
		\label{widthtest_active}
	\end{minipage}
\end{figure*}

\begin{figure*}[tb]
		\begin{minipage}[t]{0.475\linewidth}\vspace{0mm}%
		\centering
		\includegraphics[width=7cm]{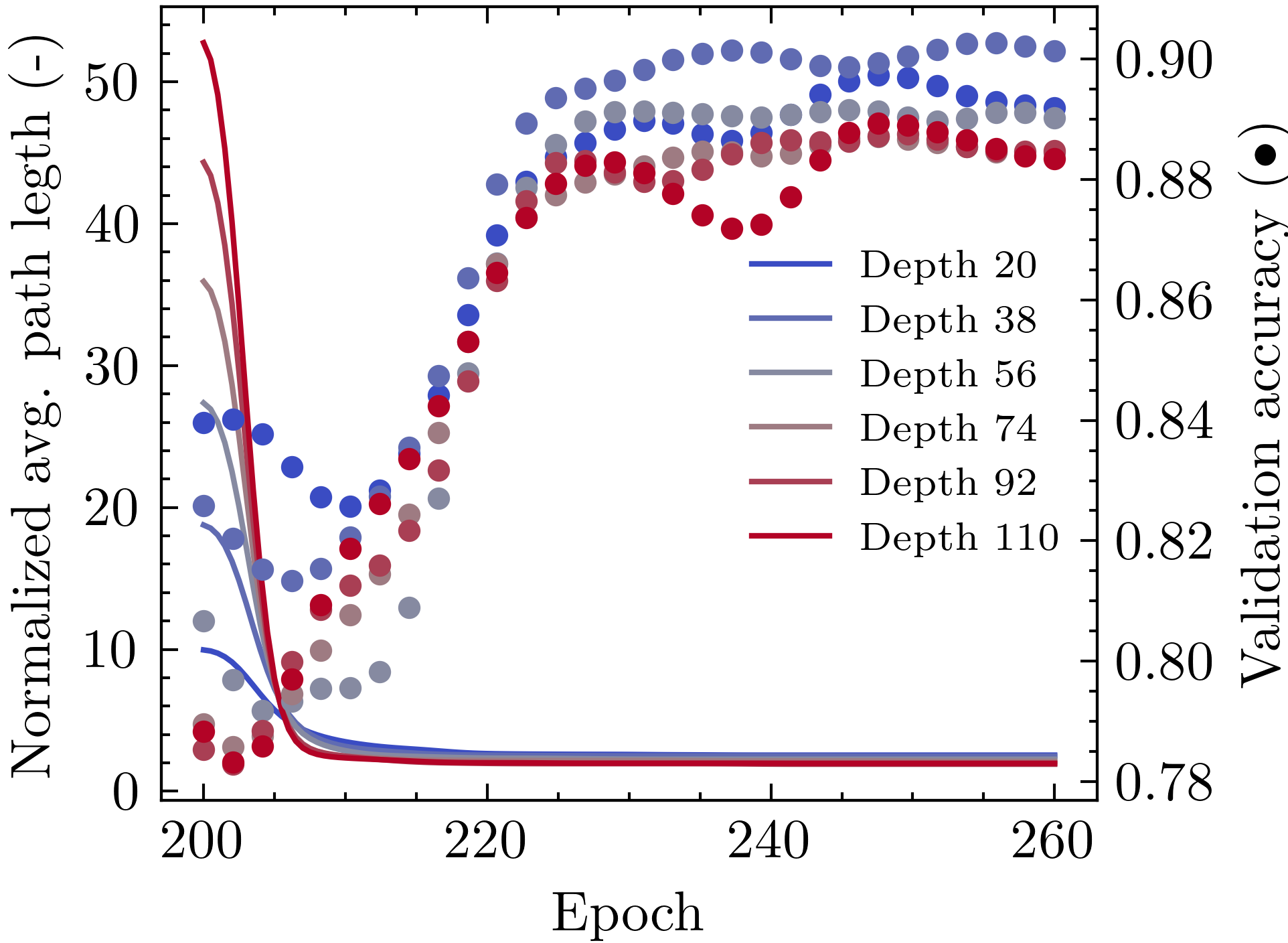}
		\caption{Validation accuracy and average path length during linearization of ResNets of different depth  for $\omega = 0.003$.}
		\label{depthtest}
	\end{minipage}
	\hfill
	\begin{minipage}[t]{0.475\linewidth}\vspace{0mm}%
		\centering
		\includegraphics[width=7cm]{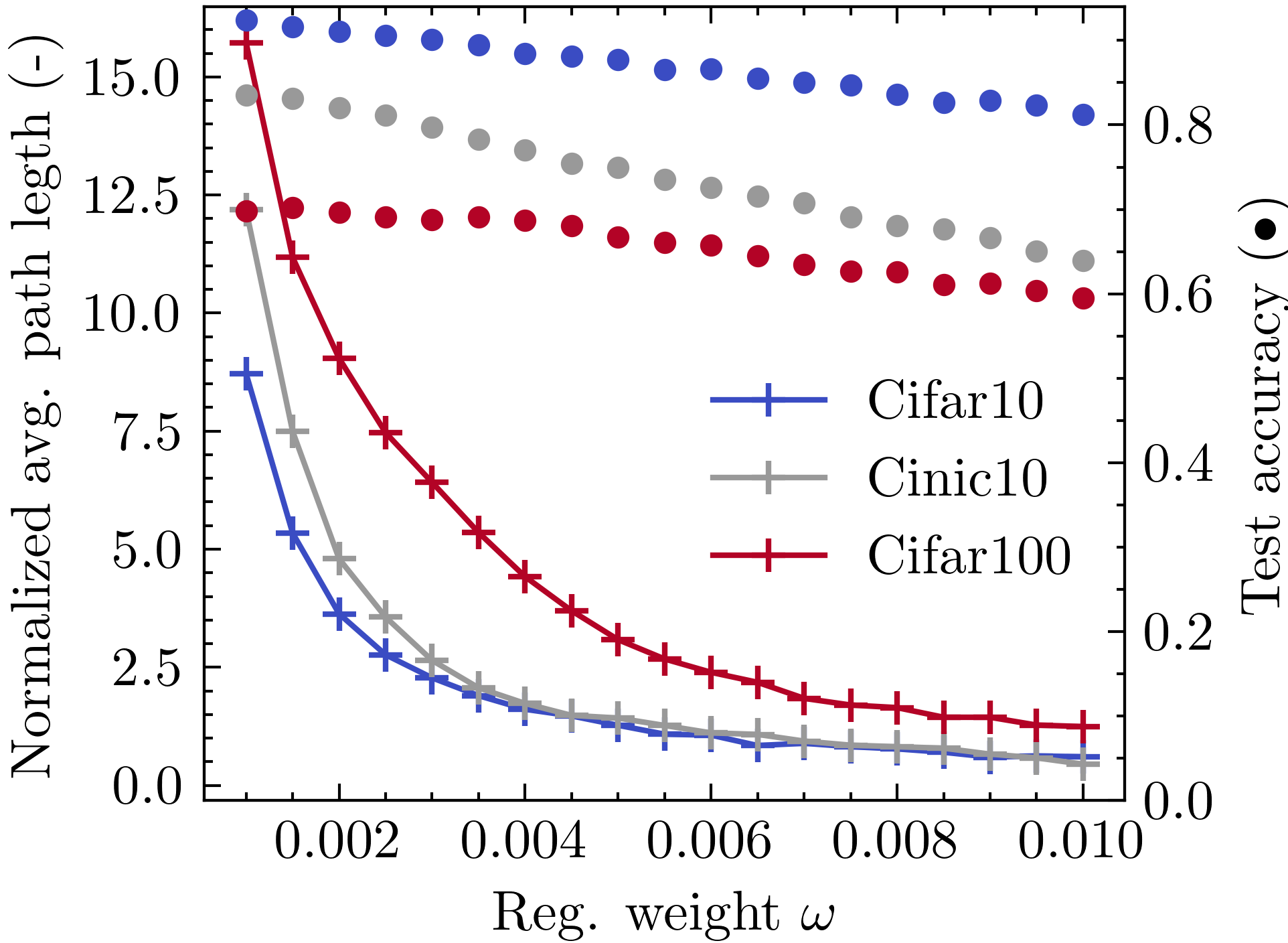}
		\caption{Test accuracy and average path length after linearization of ResNet56 for different regularization weight choices $\omega$ on the Cifar10, CINIC-10 and CIFAR-100 dataset.}
		\label{regtest_compare}
	\end{minipage}
\end{figure*}{}

\subsubsection{ENW Converges Approx. Independently of Initial Width}
\label{section_width}

We now want to analyze the effect of network width on the shape of the resulting partially linearized network. We therefore apply our linearization technique to different versions of ResNet56 Short scaled in width by a factor of $\nu\in\{0.25,0.5, 1, 1.5, 2, 2.38 \}$ and measure the NAPL and performance of the resulting networks. In Figure \ref{widthtest}, we see that wider networks seem to have better performance but lower NAPL. The relationship between the number of filters in a layer and the average path length seems reciprocal: this would imply that the average number of active neurons per layer remains approximately equal. To confirm this, in Figure \ref{widthtest_active} we plotted the inverse proportion of active PReLUs after post-training linearization for all networks. We can see a linear relationship, confirming that the average amount of active neurons per layer remains roughly constant  - independently of the initial width chosen.

\subsubsection{NAPL Converges Approx. Independently of Initial Depth}
\label{depthtest_sect}
We saw that independently of the network width chosen, there was a similar amount of neurons per layer that remained active. We now want to establish if we can make a similar statement with regard to network depth. Therefore, we repeated the experiment of the last section for networks of different depth. In Figure \ref{depthtest}, we see that independently of the initial network chosen, the resulting network's NAPL converges to a similar value while having comparable training performances. Shallower networks seemingly converge to a marginally higher NAPL but this is merely an artifact of how we scaled ResNet blocks and is not observable on a simple convolutional network with constant width (ref. Appendix Figure \ref{depthtest_ctw_toynet}). The results hint the existence of a core nonlinear structure that forms during training that is necessary to learn a given task that is approximately constant in depth and width, regardless of the initial network with chosen. Similar experiments on depth and width on CIFAR-100 in Figure \ref{depthtest_cifar100} in the Appendix show qualitatively the same behaviour but with different NAPL values.

Note that our method yields better performances than \citep{layer_folding} for a lower effective depth. The authors obtain a performance of 90.29\% / 67.04\% for a ResNet56 reduced to 10 layers (that would correspond to a NAPL close to 5) on Cifar10 / Cifar100. We obtained 90.59\% / 67.64\% for a ResNet56 with NAPL 2.7 / 3.1.

\begin{figure*}[tbp!]
	\centering
	\includegraphics[width=\linewidth]{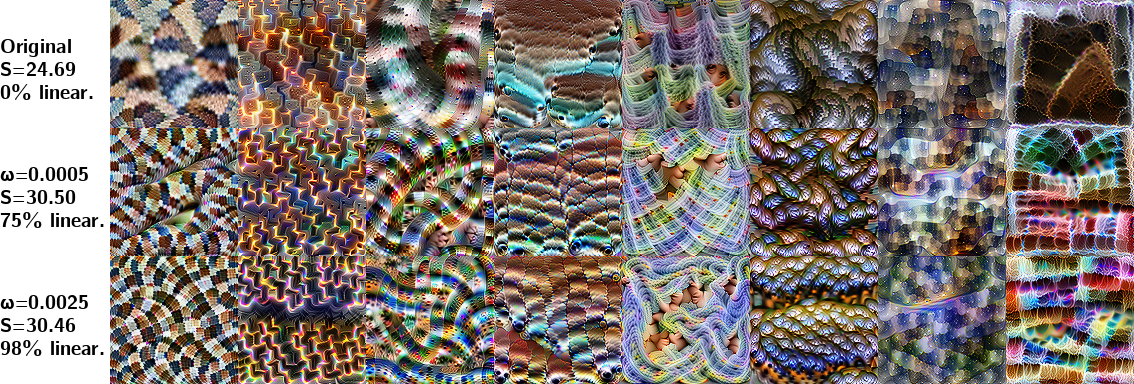}
	\caption{Visualization of \textit{the same} feature channel per column for a ResNet50 trained on ImageNet across different degrees of partial linearization. The value of $S$ indicates the average image sharpness, measured by the average image gradient magnitude.}
	\label{featureviz}
\end{figure*}

\subsubsection{NAPL Depends on Task Difficulty}
To understand how the difficulty of the task to learn affects the shape of the resulting partially linearized networks, we regularized many instances of ResNet56 Short for different choices of $\omega$ on CIFAR-10, CINIC-10 and CIFAR-100. Looking at Figure \ref{regtest_compare}, we first note that by increasing the regularization weight, the NAPL of the resulting network is decreased as expected, but this effect seems to saturate exponentially while the test accuracy is only reduced linearly in $\omega$.  This seems to imply that there is a minimum NAPL necessary to learn a given task. We also see that the NAPL measured is consistently higher for harder datasets (datasets where the networks reach a low top-1 accuracy), except for very high regularization values where the CINIC-10 and CIFAR-10 curves converge. We conclude that our method is able to extract a network with minimal nonlinearity able to learn a given task.

\subsection{Impact on Feature Visualizations}

Finally, we visualized some feature channels from ResNet50 Short trained on ImageNet and partly linearized to different degrees ($\omega\in[0,0.0005,0.0025]$) with the Lucent \citep{lucent} library; the performance and amount of remaining active PReLU units in these networks is shown in Figure \ref{performance_imgnet}. We only consider feature channels that are still active in all three networks and chose one of the deeper layers for visualization in Figure \ref{featureviz}. We see that in most cases, we can recognize a given feature over the different networks and the image seems subjectively sharper for the partially linearized networks. To confirm this, we measured the magnitude of the image gradient, averaged over all 1721 feature channels remaining in all three network and rgb-channels. This sharpness measure (indicated as $S$) is indeed significantly higher for the linearized networks, but we cannot discern a further increase for the network with higher degree of linearization.

\section{Summary of Contributions and Discussion}
\new{In our work, we observe that linearized networks extracted \textit{after} training outperform networks with the same amount of nonlinear units trained \textit{from scratch} in experiments on convolutional and transformer architectures  trained on computer vision and machine translation tasks. This is a highly surprising observation, as the final network architecture and thus it's expressivity are the same, but still, worse minima are found when training shallow architectures from scratch as opposed to training a deeper architecture and making it shallower afterwards. This is, to our best knowledge, the first time such an effect is described in literature. Complementary findings in recent literature describe a similar effect of finding a simpler "core structure"  contained in the network in early training: for fully linearizing the network with regard to its weights (time/data-dependent NTK) \cite{nonlin_advantage} and for drawing "lottery tickets" \citep{eb_lottery}. By reducing nonlinearity at channel-level, our method is able to extract networks containing significantly less nonlinear units while maintaining a similar performance, compared to previous attempts \citep{layer_folding}. We conclude that theoretical results analyzing very shallow networks e.g. \cite{linear_regions} might have higher significance on networks used in practice than previously thought and that network depth mostly benefits the training process in early stages, as most of their nonlinear expressive power or depth is not utilized after training.}

Since our method prunes nonlinearity at a channel-level as opposed to network weights, it is easier to relate the reduction in complexity to function space \new{in terms of effective width and depth}. In order to sensibly characterize the depth of a network with many linearized channels, we introduce the average path length of a network, a measure that counts the average number of nonlinear units encountered over all paths in the computation graph of the network. Linearizing networks to different degrees then allows us to characterize the minimum effective width and depth needed in order to solve a given problem. In experiments on multiple computer vision datasets, we found that these values are roughly constant for a given problem and fixed regularization weight, independently of the initial network chosen and that that the effective depth of a network grows with problem complexity. Further, we found that the essential nonlinear units in a network are distributed rather uniformly over the layers for residual networks as opposed to plain feedforward networks where they seem form clusters. This indicates that in deep networks without residual connections, there could be large connected blocks of layers that contribute very little to the learned function's nonlinear complexity.

Finally, since our method drastically reduces the amount of nonlinear feature channels in a network, we envision it can be useful for researchers trying to explain a network's behavior through visual inspection of its features. Our method not only reduces the number of nonlinear feature channels but also measurably increases the sharpness of gradient-based feature visualizations.

\section{Acknowledgments}
The authors acknowledge funding from the Emergent AI Center funded by the Carl-Zeiss-Stiftung. The authors would like to thank Daniel Franzen and Michael Wand for their helpful discussions.
\clearpage

\bibliography{references}

\begin{thebibliography}{38}
\providecommand{\natexlab}[1]{#1}
\providecommand{\url}[1]{\texttt{#1}}
\expandafter\ifx\csname urlstyle\endcsname\relax
  \providecommand{\doi}[1]{doi: #1}\else
  \providecommand{\doi}{doi: \begingroup \urlstyle{rm}\Url}\fi

\bibitem[Ali{ }Mehmeti{-}G{\"{o}}pel et~al.(2021)Ali{ }Mehmeti{-}G{\"{o}}pel,
  Hartmann, and Wand]{ringing_relus}
Ali{ }Mehmeti{-}G{\"{o}}pel, C. H.~X., Hartmann, D., and Wand, M.
\newblock Ringing relus: Harmonic distortion analysis of nonlinear feedforward
  networks.
\newblock In \emph{9th International Conference on Learning Representations,
  {ICLR} 2021, Virtual Event, Austria, May 3-7, 2021}. OpenReview.net, 2021.
\newblock URL \url{https://openreview.net/forum?id=TaYhv-q1Xit}.

\bibitem[Allen{-}Zhu et~al.(2019)Allen{-}Zhu, Li, and Liang]{generalization}
Allen{-}Zhu, Z., Li, Y., and Liang, Y.
\newblock Learning and generalization in overparameterized neural networks,
  going beyond two layers.
\newblock In Wallach, H.~M., Larochelle, H., Beygelzimer, A.,
  d'Alch{\'{e}}{-}Buc, F., Fox, E.~B., and Garnett, R. (eds.), \emph{Advances
  in Neural Information Processing Systems 32: Annual Conference on Neural
  Information Processing Systems 2019, NeurIPS 2019, December 8-14, 2019,
  Vancouver, BC, Canada}, pp.\  6155--6166, 2019.
\newblock URL
  \url{https://proceedings.neurips.cc/paper/2019/hash/62dad6e273d32235ae02b7d321578ee8-Abstract.html}.

\bibitem[Arora et~al.(2019)Arora, Li, and Lyu]{bn_autotune}
Arora, S., Li, Z., and Lyu, K.
\newblock Theoretical analysis of auto rate-tuning by batch normalization.
\newblock In \emph{7th International Conference on Learning Representations,
  {ICLR} 2019, New Orleans, LA, USA, May 6-9, 2019}. OpenReview.net, 2019.
\newblock URL \url{https://openreview.net/forum?id=rkxQ-nA9FX}.

\bibitem[Balduzzi et~al.(2017)Balduzzi, Frean, Leary, Lewis, Ma, and
  McWilliams]{shattering_gradients}
Balduzzi, D., Frean, M., Leary, L., Lewis, J.~P., Ma, K.~W., and McWilliams, B.
\newblock The shattered gradients problem: If resnets are the answer, then what
  is the question?
\newblock In Precup, D. and Teh, Y.~W. (eds.), \emph{Proceedings of the 34th
  International Conference on Machine Learning, {ICML} 2017, Sydney, NSW,
  Australia, 6-11 August 2017}, volume~70 of \emph{Proceedings of Machine
  Learning Research}, pp.\  342--350. {PMLR}, 2017.
\newblock URL \url{http://proceedings.mlr.press/v70/balduzzi17b.html}.

\bibitem[Bartlett et~al.(2019)Bartlett, Harvey, Liaw, and Mehrabian]{vc_bounds}
Bartlett, P.~L., Harvey, N., Liaw, C., and Mehrabian, A.
\newblock Nearly-tight vc-dimension and pseudodimension bounds for piecewise
  linear neural networks.
\newblock \emph{J. Mach. Learn. Res.}, 20:\penalty0 63:1--63:17, 2019.
\newblock URL \url{http://jmlr.org/papers/v20/17-612.html}.

\bibitem[Belkin et~al.(2019)Belkin, Hsu, Ma, and Mandal]{belkin2019reconciling}
Belkin, M., Hsu, D., Ma, S., and Mandal, S.
\newblock Reconciling modern machine-learning practice and the classical
  bias--variance trade-off.
\newblock \emph{Proceedings of the National Academy of Sciences}, 116\penalty0
  (32):\penalty0 15849--15854, 2019.

\bibitem[Cybenko(1989)]{original_uat}
Cybenko, G.
\newblock Approximation by superpositions of a sigmoidal function.
\newblock \emph{Math. Control. Signals Syst.}, 2\penalty0 (4):\penalty0
  303--314, 1989.
\newblock \doi{10.1007/BF02551274}.
\newblock URL \url{https://doi.org/10.1007/BF02551274}.

\bibitem[Darlow et~al.(2018)Darlow, Crowley, Antoniou, and Storkey]{cinic10}
Darlow, L.~N., Crowley, E.~J., Antoniou, A., and Storkey, A.~J.
\newblock {CINIC-10} is not imagenet or {CIFAR-10}.
\newblock \emph{CoRR}, abs/1810.03505, 2018.
\newblock URL \url{http://arxiv.org/abs/1810.03505}.

\bibitem[Deng et~al.(2009)Deng, Dong, Socher, Li, Li, and Fei{-}Fei]{imagenet}
Deng, J., Dong, W., Socher, R., Li, L., Li, K., and Fei{-}Fei, L.
\newblock Imagenet: {A} large-scale hierarchical image database.
\newblock In \emph{2009 {IEEE} Computer Society Conference on Computer Vision
  and Pattern Recognition {(CVPR} 2009), 20-25 June 2009, Miami, Florida,
  {USA}}, pp.\  248--255. {IEEE} Computer Society, 2009.
\newblock \doi{10.1109/CVPR.2009.5206848}.
\newblock URL \url{https://doi.org/10.1109/CVPR.2009.5206848}.

\bibitem[Dror et~al.(2021)Dror, Zehngut, Raviv, Artyomov, Vitek, and
  Jevnisek]{layer_folding}
Dror, A.~B., Zehngut, N., Raviv, A., Artyomov, E., Vitek, R., and Jevnisek,
  R.~J.
\newblock Layer folding: Neural network depth reduction using activation
  linearization.
\newblock \emph{CoRR}, abs/2106.09309, 2021.
\newblock URL \url{https://arxiv.org/abs/2106.09309}.

\bibitem[Elliott et~al.(2016)Elliott, Frank, Sima'an, and Specia]{multi30k}
Elliott, D., Frank, S., Sima'an, K., and Specia, L.
\newblock Multi30k: Multilingual english-german image descriptions.
\newblock In \emph{Proceedings of the 5th Workshop on Vision and Language,
  hosted by the 54th Annual Meeting of the Association for Computational
  Linguistics, VL@ACL 2016, August 12, Berlin, Germany}. The Association for
  Computer Linguistics, 2016.
\newblock \doi{10.18653/v1/w16-3210}.
\newblock URL \url{https://doi.org/10.18653/v1/w16-3210}.

\bibitem[Fort et~al.(2020)Fort, Dziugaite, Paul, Kharaghani, Roy, and
  Ganguli]{nonlin_advantage}
Fort, S., Dziugaite, G.~K., Paul, M., Kharaghani, S., Roy, D.~M., and Ganguli,
  S.
\newblock Deep learning versus kernel learning: an empirical study of loss
  landscape geometry and the time evolution of the neural tangent kernel.
\newblock In Larochelle, H., Ranzato, M., Hadsell, R., Balcan, M., and Lin, H.
  (eds.), \emph{Advances in Neural Information Processing Systems 33: Annual
  Conference on Neural Information Processing Systems 2020, NeurIPS 2020,
  December 6-12, 2020, virtual}, 2020.
\newblock URL
  \url{https://proceedings.neurips.cc/paper/2020/hash/405075699f065e43581f27d67bb68478-Abstract.html}.

\bibitem[Frankle \& Carbin(2019)Frankle and Carbin]{lottery}
Frankle, J. and Carbin, M.
\newblock The lottery ticket hypothesis: Finding sparse, trainable neural
  networks.
\newblock In \emph{7th International Conference on Learning Representations,
  {ICLR} 2019, New Orleans, LA, USA, May 6-9, 2019}. OpenReview.net, 2019.
\newblock URL \url{https://openreview.net/forum?id=rJl-b3RcF7}.

\bibitem[Gebhart et~al.(2021)Gebhart, Saxena, and Schrater]{pruning_init}
Gebhart, T., Saxena, U., and Schrater, P.
\newblock A unified paths perspective for pruning at initialization.
\newblock \emph{CoRR}, abs/2101.10552, 2021.
\newblock URL \url{https://arxiv.org/abs/2101.10552}.

\bibitem[Han et~al.(2017)Han, Kim, and Kim]{pyramidnet}
Han, D., Kim, J., and Kim, J.
\newblock Deep pyramidal residual networks.
\newblock In \emph{2017 {IEEE} Conference on Computer Vision and Pattern
  Recognition, {CVPR} 2017, Honolulu, HI, USA, July 21-26, 2017}, pp.\
  6307--6315. {IEEE} Computer Society, 2017.
\newblock \doi{10.1109/CVPR.2017.668}.
\newblock URL \url{https://doi.org/10.1109/CVPR.2017.668}.

\bibitem[Hanin \& Rolnick(2019)Hanin and Rolnick]{few_patterns}
Hanin, B. and Rolnick, D.
\newblock Deep relu networks have surprisingly few activation patterns.
\newblock In Wallach, H.~M., Larochelle, H., Beygelzimer, A.,
  d'Alch{\'{e}}{-}Buc, F., Fox, E.~B., and Garnett, R. (eds.), \emph{Advances
  in Neural Information Processing Systems 32: Annual Conference on Neural
  Information Processing Systems 2019, NeurIPS 2019, December 8-14, 2019,
  Vancouver, BC, Canada}, pp.\  359--368, 2019.
\newblock URL
  \url{https://proceedings.neurips.cc/paper/2019/hash/9766527f2b5d3e95d4a733fcfb77bd7e-Abstract.html}.

\bibitem[He et~al.(2015{\natexlab{a}})He, Zhang, Ren, and Sun]{prelu}
He, K., Zhang, X., Ren, S., and Sun, J.
\newblock Delving deep into rectifiers: Surpassing human-level performance on
  imagenet classification.
\newblock In \emph{2015 {IEEE} International Conference on Computer Vision,
  {ICCV} 2015, Santiago, Chile, December 7-13, 2015}, pp.\  1026--1034. {IEEE}
  Computer Society, 2015{\natexlab{a}}.
\newblock \doi{10.1109/ICCV.2015.123}.
\newblock URL \url{https://doi.org/10.1109/ICCV.2015.123}.

\bibitem[He et~al.(2015{\natexlab{b}})He, Zhang, Ren, and Sun]{resnet}
He, K., Zhang, X., Ren, S., and Sun, J.
\newblock Delving deep into rectifiers: Surpassing human-level performance on
  imagenet classification.
\newblock In \emph{2015 {IEEE} International Conference on Computer Vision,
  {ICCV} 2015, Santiago, Chile, December 7-13, 2015}, pp.\  1026--1034. {IEEE}
  Computer Society, 2015{\natexlab{b}}.
\newblock \doi{10.1109/ICCV.2015.123}.
\newblock URL \url{https://doi.org/10.1109/ICCV.2015.123}.

\bibitem[He et~al.(2016)He, Zhang, Ren, and Sun]{resnetv2}
He, K., Zhang, X., Ren, S., and Sun, J.
\newblock Identity mappings in deep residual networks.
\newblock In Leibe, B., Matas, J., Sebe, N., and Welling, M. (eds.),
  \emph{Computer Vision - {ECCV} 2016 - 14th European Conference, Amsterdam,
  The Netherlands, October 11-14, 2016, Proceedings, Part {IV}}, volume 9908 of
  \emph{Lecture Notes in Computer Science}, pp.\  630--645. Springer, 2016.
\newblock \doi{10.1007/978-3-319-46493-0\_38}.

\bibitem[Hochreiter(1991)]{vanishing_gradients}
Hochreiter, S.
\newblock Untersuchungen zu dynamischen neuronalen netzen.
\newblock Diploma thesis, TU Munich, 1991.

\bibitem[Ioffe \& Szegedy(2015)Ioffe and Szegedy]{batchnorm}
Ioffe, S. and Szegedy, C.
\newblock Batch normalization: Accelerating deep network training by reducing
  internal covariate shift.
\newblock In Bach, F.~R. and Blei, D.~M. (eds.), \emph{Proceedings of the 32nd
  International Conference on Machine Learning, {ICML} 2015, Lille, France,
  6-11 July 2015}, volume~37 of \emph{{JMLR} Workshop and Conference
  Proceedings}, pp.\  448--456. JMLR.org, 2015.

\bibitem[Jacot et~al.(2018)Jacot, Gabriel, and Hongler]{Jacot2018}
Jacot, A., Gabriel, F., and Hongler, C.
\newblock Neural tangent kernel: Convergence and generalization in neural
  networks.
\newblock volume~31, pp.\  8571--8580. Curran Associates, Inc., 2018.

\bibitem[Kiat(2021)]{lucent}
Kiat, L.~S.
\newblock Lucent.
\newblock \url{https://github.com/greentfrapp/lucent}, 2021.

\bibitem[Krizhevsky(2009)]{cifar}
Krizhevsky, A.
\newblock Learning multiple layers of features from tiny images.
\newblock pp.\  32--33, 2009.
\newblock URL
  \url{https://www.cs.toronto.edu/~kriz/learning-features-2009-TR.pdf}.

\bibitem[Krizhevsky et~al.(2012)Krizhevsky, Sutskever, and Hinton]{alexnet}
Krizhevsky, A., Sutskever, I., and Hinton, G.~E.
\newblock Imagenet classification with deep convolutional neural networks.
\newblock In Bartlett, P.~L., Pereira, F. C.~N., Burges, C. J.~C., Bottou, L.,
  and Weinberger, K.~Q. (eds.), \emph{Advances in Neural Information Processing
  Systems 25: 26th Annual Conference on Neural Information Processing Systems
  2012. Proceedings of a meeting held December 3-6, 2012, Lake Tahoe, Nevada,
  United States}, pp.\  1106--1114, 2012.
\newblock URL
  \url{https://proceedings.neurips.cc/paper/2012/hash/c399862d3b9d6b76c8436e924a68c45b-Abstract.html}.

\bibitem[Lakshminarayanan \& Singh(2020)Lakshminarayanan and
  Singh]{Lakshminarayanan2020}
Lakshminarayanan, C. and Singh, A.~V.
\newblock Neural path features and neural path kernel : Understanding the role
  of gates in deep learning.
\newblock In \emph{Advances in Neural Information Processing Systems(NeurIPS)},
  volume~33, 2020.

\bibitem[Le \& Yang(2015)Le and Yang]{tiny_imagenet}
Le, Y. and Yang, X.~S.
\newblock Tiny imagenet visual recognition challenge.
\newblock 2015.

\bibitem[Neill(2020)]{compression_survey}
Neill, J.~O.
\newblock An overview of neural network compression.
\newblock \emph{arXiv preprint arXiv:2006.03669}, 2020.

\bibitem[Roberts et~al.(2022)Roberts, Yaida, and Hanin]{Roberts2022}
Roberts, D.~A., Yaida, S., and Hanin, B.
\newblock \emph{The Principles of Deep Learning Theory}.
\newblock Cambridge University Press, 2022.
\newblock \url{https://deeplearningtheory.com}.

\bibitem[Safran et~al.(2022)Safran, Vardi, and Lee]{linear_regions}
Safran, I., Vardi, G., and Lee, J.~D.
\newblock On the effective number of linear regions in shallow univariate relu
  networks: Convergence guarantees and implicit bias.
\newblock \emph{CoRR}, abs/2205.09072, 2022.
\newblock \doi{10.48550/arXiv.2205.09072}.
\newblock URL \url{https://doi.org/10.48550/arXiv.2205.09072}.

\bibitem[Santurkar et~al.(2018)Santurkar, Tsipras, Ilyas, and
  Madry]{bn_covshift}
Santurkar, S., Tsipras, D., Ilyas, A., and Madry, A.
\newblock How does batch normalization help optimization?
\newblock In Bengio, S., Wallach, H.~M., Larochelle, H., Grauman, K.,
  Cesa{-}Bianchi, N., and Garnett, R. (eds.), \emph{Advances in Neural
  Information Processing Systems 31: Annual Conference on Neural Information
  Processing Systems 2018, NeurIPS 2018, December 3-8, 2018, Montr{\'{e}}al,
  Canada}, pp.\  2488--2498, 2018.
\newblock URL
  \url{https://proceedings.neurips.cc/paper/2018/hash/905056c1ac1dad141560467e0a99e1cf-Abstract.html}.

\bibitem[Su et~al.(2020)Su, Chen, Cai, Wu, Gao, Wang, and Lee]{sanity_lottery}
Su, J., Chen, Y., Cai, T., Wu, T., Gao, R., Wang, L., and Lee, J.~D.
\newblock Sanity-checking pruning methods: Random tickets can win the jackpot.
\newblock In Larochelle, H., Ranzato, M., Hadsell, R., Balcan, M., and Lin, H.
  (eds.), \emph{Advances in Neural Information Processing Systems 33: Annual
  Conference on Neural Information Processing Systems 2020, NeurIPS 2020,
  December 6-12, 2020, virtual}, 2020.
\newblock URL
  \url{https://proceedings.neurips.cc/paper/2020/hash/eae27d77ca20db309e056e3d2dcd7d69-Abstract.html}.

\bibitem[Tan \& Le(2019)Tan and Le]{efficientnet}
Tan, M. and Le, Q.~V.
\newblock Efficientnet: Rethinking model scaling for convolutional neural
  networks.
\newblock In Chaudhuri, K. and Salakhutdinov, R. (eds.), \emph{Proceedings of
  the 36th International Conference on Machine Learning, {ICML} 2019, 9-15 June
  2019, Long Beach, California, {USA}}, volume~97 of \emph{Proceedings of
  Machine Learning Research}, pp.\  6105--6114. {PMLR}, 2019.
\newblock URL \url{http://proceedings.mlr.press/v97/tan19a.html}.

\bibitem[Vaswani et~al.(2017)Vaswani, Shazeer, Parmar, Uszkoreit, Jones, Gomez,
  Kaiser, and Polosukhin]{transformer}
Vaswani, A., Shazeer, N., Parmar, N., Uszkoreit, J., Jones, L., Gomez, A.~N.,
  Kaiser, L., and Polosukhin, I.
\newblock Attention is all you need.
\newblock In Guyon, I., von Luxburg, U., Bengio, S., Wallach, H.~M., Fergus,
  R., Vishwanathan, S. V.~N., and Garnett, R. (eds.), \emph{Advances in Neural
  Information Processing Systems 30: Annual Conference on Neural Information
  Processing Systems 2017, December 4-9, 2017, Long Beach, CA, {USA}}, pp.\
  5998--6008, 2017.
\newblock URL
  \url{https://proceedings.neurips.cc/paper/2017/hash/3f5ee243547dee91fbd053c1c4a845aa-Abstract.html}.

\bibitem[Veit et~al.(2016)Veit, Wilber, and Belongie]{resnet_ensemble}
Veit, A., Wilber, M.~J., and Belongie, S.~J.
\newblock Residual networks behave like ensembles of relatively shallow
  networks.
\newblock In Lee, D.~D., Sugiyama, M., von Luxburg, U., Guyon, I., and Garnett,
  R. (eds.), \emph{Advances in Neural Information Processing Systems 29: Annual
  Conference on Neural Information Processing Systems 2016, December 5-10,
  2016, Barcelona, Spain}, pp.\  550--558, 2016.
\newblock URL
  \url{https://proceedings.neurips.cc/paper/2016/hash/37bc2f75bf1bcfe8450a1a41c200364c-Abstract.html}.

\bibitem[Wilson \& Izmailov(2020)Wilson and Izmailov]{wilson2020bayesian}
Wilson, A.~G. and Izmailov, P.
\newblock Bayesian deep learning and a probabilistic perspective of
  generalization.
\newblock \emph{Advances in neural information processing systems},
  33:\penalty0 4697--4708, 2020.

\bibitem[You et~al.(2020)You, Li, Xu, Fu, Wang, Chen, Baraniuk, Wang, and
  Lin]{eb_lottery}
You, H., Li, C., Xu, P., Fu, Y., Wang, Y., Chen, X., Baraniuk, R.~G., Wang, Z.,
  and Lin, Y.
\newblock Drawing early-bird tickets: Toward more efficient training of deep
  networks.
\newblock In \emph{8th International Conference on Learning Representations,
  {ICLR} 2020, Addis Ababa, Ethiopia, April 26-30, 2020}. OpenReview.net, 2020.
\newblock URL \url{https://openreview.net/forum?id=BJxsrgStvr}.

\bibitem[Zhang et~al.(2019)Zhang, Dauphin, and Ma]{fixup}
Zhang, H., Dauphin, Y.~N., and Ma, T.
\newblock Fixup initialization: Residual learning without normalization.
\newblock In \emph{7th International Conference on Learning Representations,
  {ICLR} 2019, New Orleans, LA, USA, May 6-9, 2019}. OpenReview.net, 2019.
\newblock URL \url{https://openreview.net/forum?id=H1gsz30cKX}.

\end{thebibliography}
\bibliographystyle{icml2023} 
\newpage
\appendix
\onecolumn
\section{Details About Histogram Computation}
\label{hist_appendix}

\begin{figure*}[b]
	\resizebox{0.8\textwidth}{!}{
		\begin{tikzpicture}[
	> = stealth, 
	shorten > = 1pt, 
	auto,
	node distance = 3cm, 
	semithick 
	]
	
	\tikzstyle{every state}=[
	draw = black,
	thick,
	fill = white,
	minimum size = 4mm
	]
	
	\tikzstyle{relu}=[
	shape=circle,
	draw = red,
	thick,
	fill = white,
	minimum size = 4mm
	]

	\node[state] (i) {$1/0/0$};
	\node[state] (l11) [above of=i] {$1/0/0$};
	\node[relu] (l12) [left of=l11] {$0/1/0$};
	\node[state] (l13) [right of=l11] {$1/0/0$};
	
	\node[relu] (l21) [above of=l11] {$0/1/0$};
	\node[state] (l22) [left of=l21] {$2/1/0$};
	\node[relu] (l23) [right of=l21] {$0/2/0$};
	
	\node[relu] (l31) [above of=l21] {$0/2/4$};
	\node[state] (l32) [left of=l31] {$0/1/0$};
	\node[state] (l33) [right of=l31] {$0/2/0$};
	
	\node[state] (l41) [above of=l31] {$0/5/4$};
	
	\path[->] (i) edge node {} (l11);
	\path[->] (i) edge node {} (l12);
	\path[->] (i) edge node {} (l13);
	\path[->] (l11) edge node {} (l21);
	\path[->] (l11) edge node {} (l22);
	\path[->] (l11) edge node {} (l23);
	\path[->] (l12) edge node {} (l22);
	\path[->] (l13) edge node {} (l22);
	\path[->] (l13) edge node {} (l23);
	\path[->] (l21) edge node {} (l31);
	\path[->] (l21) edge node {} (l32);
	\path[->] (l22) edge node {} (l31);
	\path[->] (l23) edge node {} (l31);
	\path[->] (l23) edge node {} (l33);
	\path[->] (l31) edge node {} (l41);
	\path[->] (l32) edge node {} (l41);
	\path[->] (l33) edge node {} (l41);
\end{tikzpicture}
	\hspace{5cm}
		\begin{tikzpicture}[
	> = stealth, 
	shorten > = 1pt, 
	auto,
	node distance = 3cm, 
	semithick 
	]
	
	\tikzstyle{every state}=[
	draw = black,
	thick,
	fill = white,
	minimum size = 4mm
	]
	
	\tikzstyle{relu}=[
	shape=circle,
	draw = red,
	thick,
	fill = white,
	minimum size = 4mm
	]

	\node[state] (i) {$1/0/0$};
	\node[state] (l11) [above of=i] {$1/0/0$};
	\node[relu] (l12) [left of=l11] {$0/1/0$};
	\node[state] (l13) [right of=l11] {$1/0/0$};
	
	\node[relu] (l21) [above of=l11] {$0/1/0$};
	\node[state] (l22) [left of=l21] {$2/1/0$};
	\node[relu] (l23) [right of=l21] {$0/2/0$};
	
	\node[relu] (l31) [above of=l21] {$0/3/4$};
	\node[state] (l32) [left of=l31] {$0/2/0$};
	\node[state] (l33) [right of=l31] {$1/2/0$};
	
	\node[state] (l41) [above of=l31] {$0/7/4$};
	
	\path[->] (i) edge node {} (l11);
	\path[->] (i) edge node {} (l12);
	\path[->] (i) edge node {} (l13);
	\path[->] (l11) edge node {} (l21);
	\path[->] (l11) edge node {} (l22);
	\path[->] (l11) edge node {} (l23);
	\path[->] (l12) edge node {} (l22);
	\path[->] (l13) edge node {} (l22);
	\path[->] (l13) edge node {} (l23);
	\path[->] (l21) edge node {} (l31);
	\path[->] (l21) edge node {} (l32);
	\path[->] (l22) edge node {} (l31);
	\path[->] (l23) edge node {} (l31);
	\path[->] (l23) edge node {} (l33);
	\path[->] (l31) edge node {} (l41);
	\path[->] (l32) edge node {} (l41);
	\path[->] (l33) edge node {} (l41);
	
		\path[->] (l13) edge [bend right,  color=blue] node {} (l33);
		\path[->] (l12) edge [bend left, color=blue] node {} (l32);
		\path[->] (l11) edge [bend right, color=blue] node {} (l31);
\end{tikzpicture}}
	\caption{Computing the histogram of path lengths through dynamic programming for a non-residual (left) and a residual (right) network. Red circles designate nodes with an active PReLU activation, blue edges designate residual connections.}
	\label{hist_example}
\end{figure*}
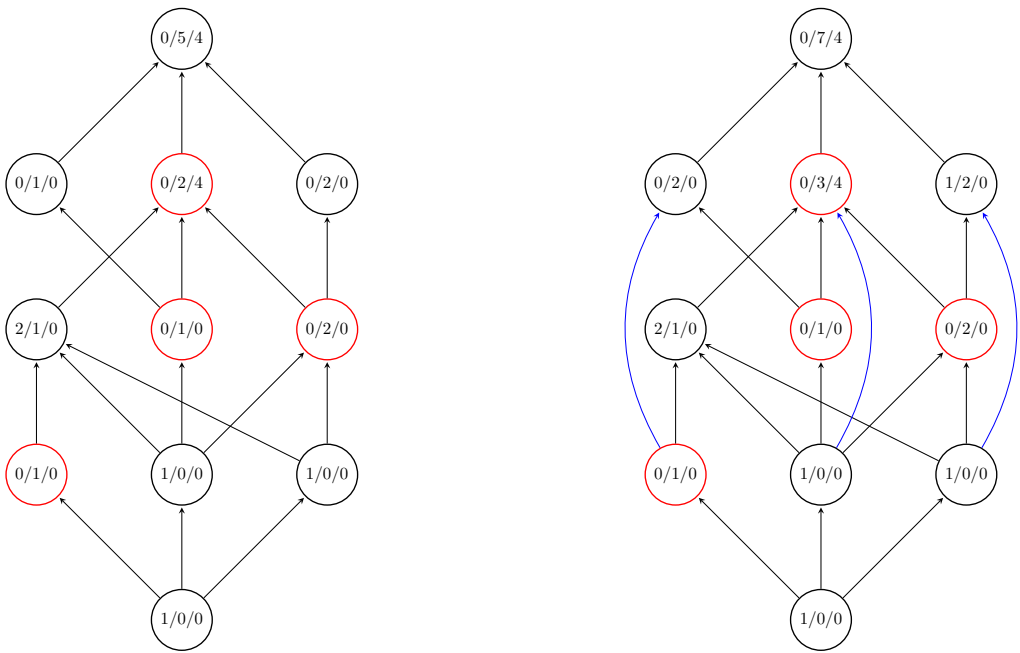

\subsection{Normalizing the Histogram}

\begin{figure*}[h]
	\resizebox{0.8\textwidth}{!}{
		\begin{tikzpicture}[
	> = stealth, 
	shorten > = 1pt, 
	auto,
	node distance = 3cm, 
	semithick 
	]
	
	\tikzstyle{every state}=[
	draw = black,
	thick,
	fill = white,
	minimum size = 4mm
	]
	
	\tikzstyle{relu}=[
	shape=circle,
	draw = red,
	thick,
	fill = white,
	minimum size = 4mm
	]

	\node[relu] (i) {$1/0/0$};
	\node[relu] (l11) [above of=i] {$0/1/0/0$};
	\node[relu] (l12) [left of=l11] {$0/1/0/0$};
	\node[relu] (l13) [right of=l11] {$0/1/0/0$};

	\node[relu] (l21) [above of=l11] {$0/0/3/0$};
	\node[relu] (l22) [left of=l21] {$0/0/3/0$};
	\node[relu] (l23) [right of=l21] {$0/0/3/0$};
	
	\node[relu] (l31) [above of=l21] {$0/0/3/9$};
	\node[relu] (l32) [left of=l31] {$0/0/3/9$};
	\node[relu] (l33) [right of=l31] {$0/0/3/9$};
	
	\path[->] (i) edge node {} (l11);
	\path[->] (i) edge node {} (l12);
	\path[->] (i) edge node {} (l13);
	\path[->] (l11) edge node {} (l21);
	\path[->] (l11) edge node {} (l22);
	\path[->] (l11) edge node {} (l23);
	\path[->] (l12) edge node {} (l21);
	\path[->] (l12) edge node {} (l22);
	\path[->] (l12) edge node {} (l23);
	\path[->] (l13) edge node {} (l21);
	\path[->] (l13) edge node {} (l22);
	\path[->] (l13) edge node {} (l23);
	\path[->] (l21) edge node {} (l31);
	\path[->] (l21) edge node {} (l32);
	\path[->] (l21) edge node {} (l33);
	\path[->] (l22) edge node {} (l31);
	\path[->] (l22) edge node {} (l32);
	\path[->] (l22) edge node {} (l33);
	\path[->] (l23) edge node {} (l31);
	\path[->] (l23) edge node {} (l32);
	\path[->] (l23) edge node {} (l33);
	\path[->] (l13) edge [bend right,  color=blue] node {} (l33);
	\path[->] (l12) edge [bend left, color=blue] node {} (l32);
	\path[->] (l11) edge [bend right, color=blue] node {} (l31);
\end{tikzpicture}
	\hspace{5cm}
		\begin{tikzpicture}[
	> = stealth, 
	shorten > = 1pt, 
	auto,
	node distance = 3cm, 
	semithick 
	]
	
	\tikzstyle{every state}=[
	draw = black,
	thick,
	fill = white,
	minimum size = 4mm
	]
	
	\tikzstyle{relu}=[
	shape=circle,
	draw = red,
	thick,
	fill = white,
	minimum size = 4mm
	]

	\node[relu] (i) {$1/0/0$};
	\node[relu] (l11) [above of=i] {$0/1/0/0$};
	\node[relu] (l12) [left of=l11] {$0/1/0/0$};
	\node[relu] (l13) [right of=l11] {$0/1/0/0$};

	\node[relu] (l21) [above of=l11] {$0/0/3/0$};
	\node[relu] (l22) [left of=l21] {$0/0/3/0$};
	\node[relu] (l23) [right of=l21] {$0/0/3/0$};
	
	\node[relu] (l31) [above of=l21] {$0/0/\frac{1}{2}/\frac{1}{2}$};
	\node[relu] (l32) [left of=l31]{$0/0/\frac{1}{2}/\frac{1}{2}$};
	\node[relu] (l33) [right of=l31]{$0/0/\frac{1}{2}/\frac{1}{2}$};
	
	\path[->] (i) edge node {} (l11);
	\path[->] (i) edge node {} (l12);
	\path[->] (i) edge node {} (l13);
	\path[->] (l11) edge node {} (l21);
	\path[->] (l11) edge node {} (l22);
	\path[->] (l11) edge node {} (l23);
	\path[->] (l12) edge node {} (l21);
	\path[->] (l12) edge node {} (l22);
	\path[->] (l12) edge node {} (l23);
	\path[->] (l13) edge node {} (l21);
	\path[->] (l13) edge node {} (l22);
	\path[->] (l13) edge node {} (l23);
	\path[->] (l21) edge node {} (l31);
	\path[->] (l21) edge node {} (l32);
	\path[->] (l21) edge node {} (l33);
	\path[->] (l22) edge node {} (l31);
	\path[->] (l22) edge node {} (l32);
	\path[->] (l22) edge node {} (l33);
	\path[->] (l23) edge node {} (l31);
	\path[->] (l23) edge node {} (l32);
	\path[->] (l23) edge node {} (l33);
	\path[->] (l13) edge [bend right,  color=blue] node {} (l33);
	\path[->] (l12) edge [bend left, color=blue] node {} (l32);
	\path[->] (l11) edge [bend right, color=blue] node {} (l31);
\end{tikzpicture}}
	\caption{Computing the unnormalized (left) and normalized (right) histogram of a residual network where all PReLUs are active. Without normalization, the residual contribution vanishes.}
	\label{hist_explosion}
\end{figure*}
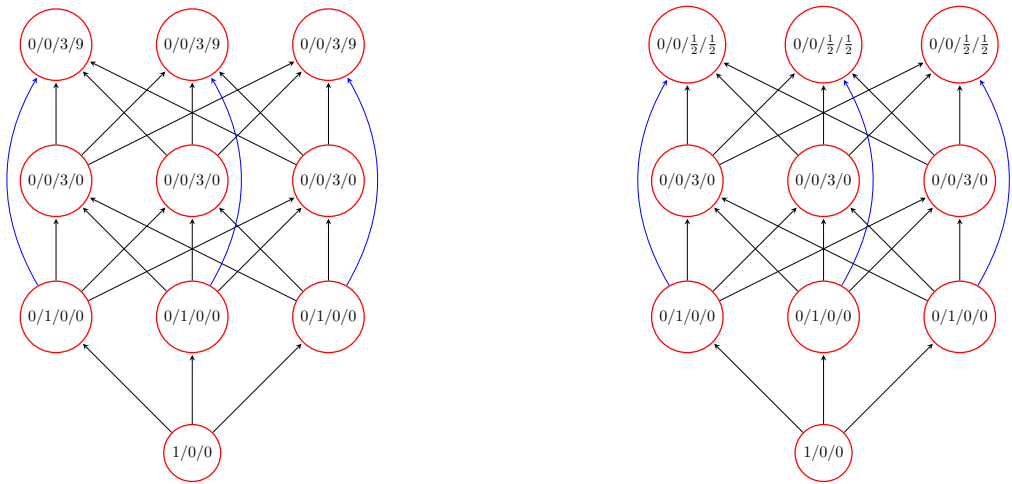

\citet{resnet_ensemble} also study the distribution of path lengths in neural networks. To model the path distribution in a residual network, the authors simply use a binomial distribution, giving the main and residual branch of a residual block equal weight. Since the authors use a binomial distribution to model path length, the average path length of a ResNet of depth $d$ before linearization is $d/2$ in their model. Figure \ref{hist_explosion} (left) makes it clear that in our (non-normalized) model as described above, the average path length would be much closer to $d$ since the main path contribution is exponentially bigger than the residual path contribution and thus vanishes in depth. 

In order to obtain an equal contribution from the main branch and the residual branch of residual block, we can modify our model to normalize the histograms before adding them together as shown in Figure \ref{hist_explosion} (right). Since we explicitly modeled the length of residual connections (how many layers are skipped), the path length of a ResNet before linearization depends on ResBlock size in our model. For a standard ResNet using BasicBlock (ResBlock size 2), we found the initial NAPL to be close to $d/2$, making our normalized average path length model similar to the one from \citet{resnet_ensemble}.

\section{Complementary Experiments}
\label{appendix:sec:further_exp}

\subsection{Plots for APL Instead of NAPL}
\begin{figure*}[tb]
	\begin{minipage}[t]{0.475\linewidth}\vspace{0mm}%
		\centering
		\includegraphics[width=7cm]{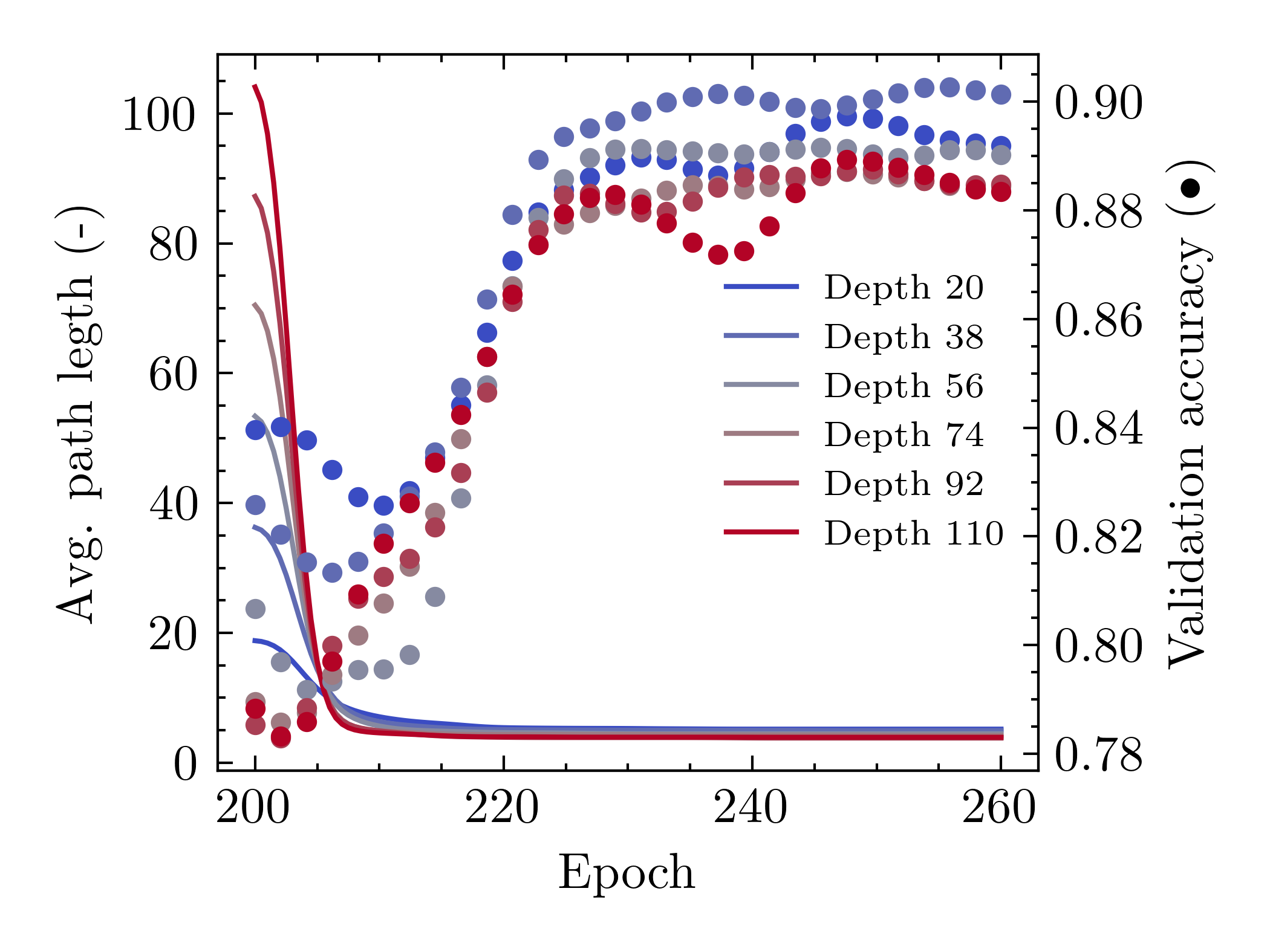}
		\caption{Validation accuracy and \textbf{unnormalized} average path length during linearization of ResNets of different depth for $\omega = 0.003$.}
		\label{depth_apl}
	\end{minipage}
	\hfill
	\begin{minipage}[t]{0.475\linewidth}\vspace{0mm}%
		\centering
		\includegraphics[width=7cm]{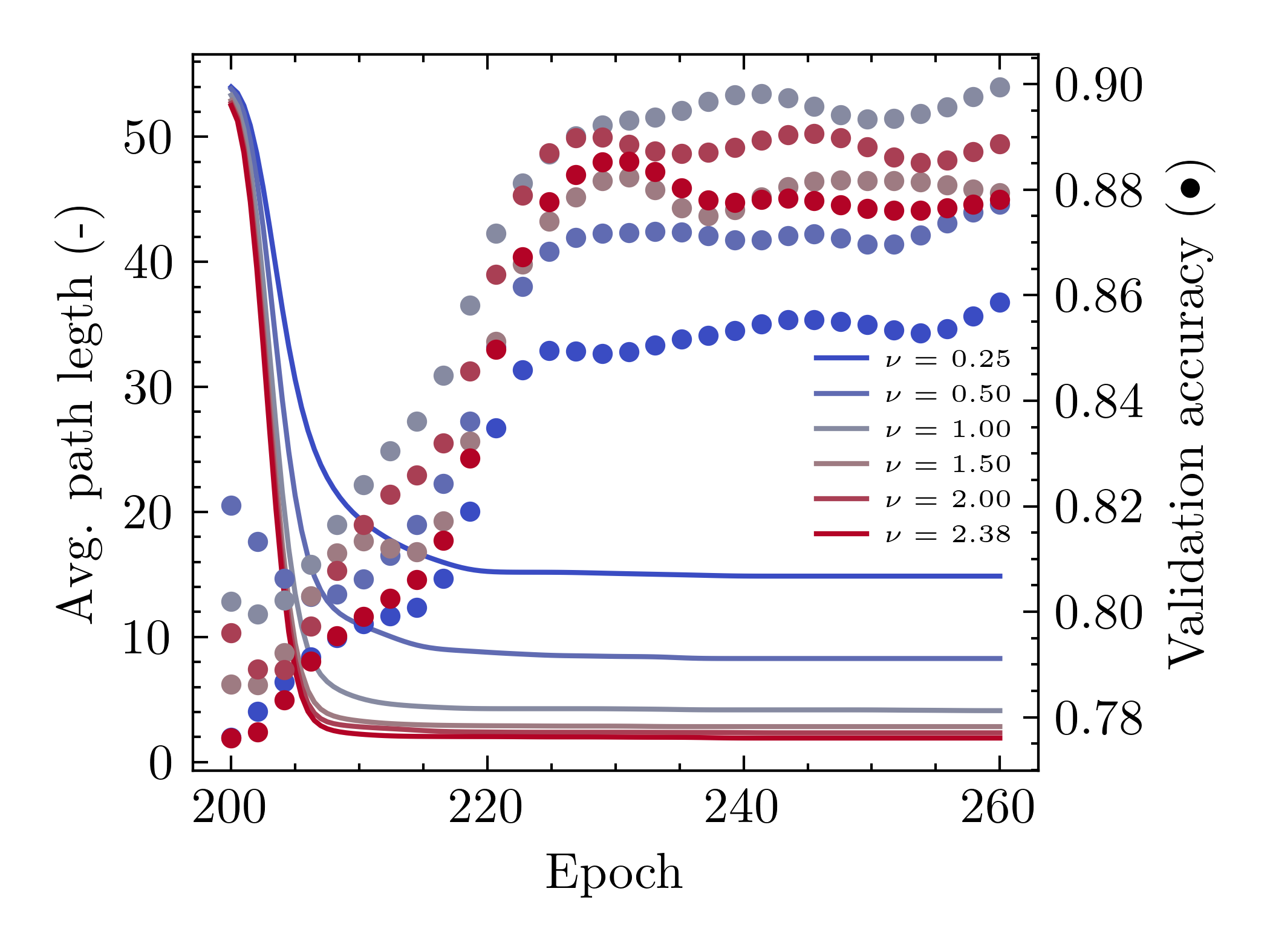}
		\caption{Validation accuracy and \textbf{unnormalized} average path length during linearization of ResNets of different width for $\omega = 0.003$.}
		\label{width_apl}
	\end{minipage}
\end{figure*}

In this section, we repeat the experiments of Figure \ref{performance}, \ref{depthtest} and \ref{widthtest} showing APL instead of NAPL. In Figure \ref{performance_apl}, \ref{depth_apl} and \ref{width_apl}, we see fundamentally no differ
ence in the results except for all unnormalized path length values being consistently higher than the normalized ones.

\subsection{Plots for Active PReLU Percentage Instead of NAPL}
\begin{figure*}[tb]
	\begin{minipage}[t]{0.475\linewidth}\vspace{0mm}%
		\centering
		\includegraphics[width=7cm]{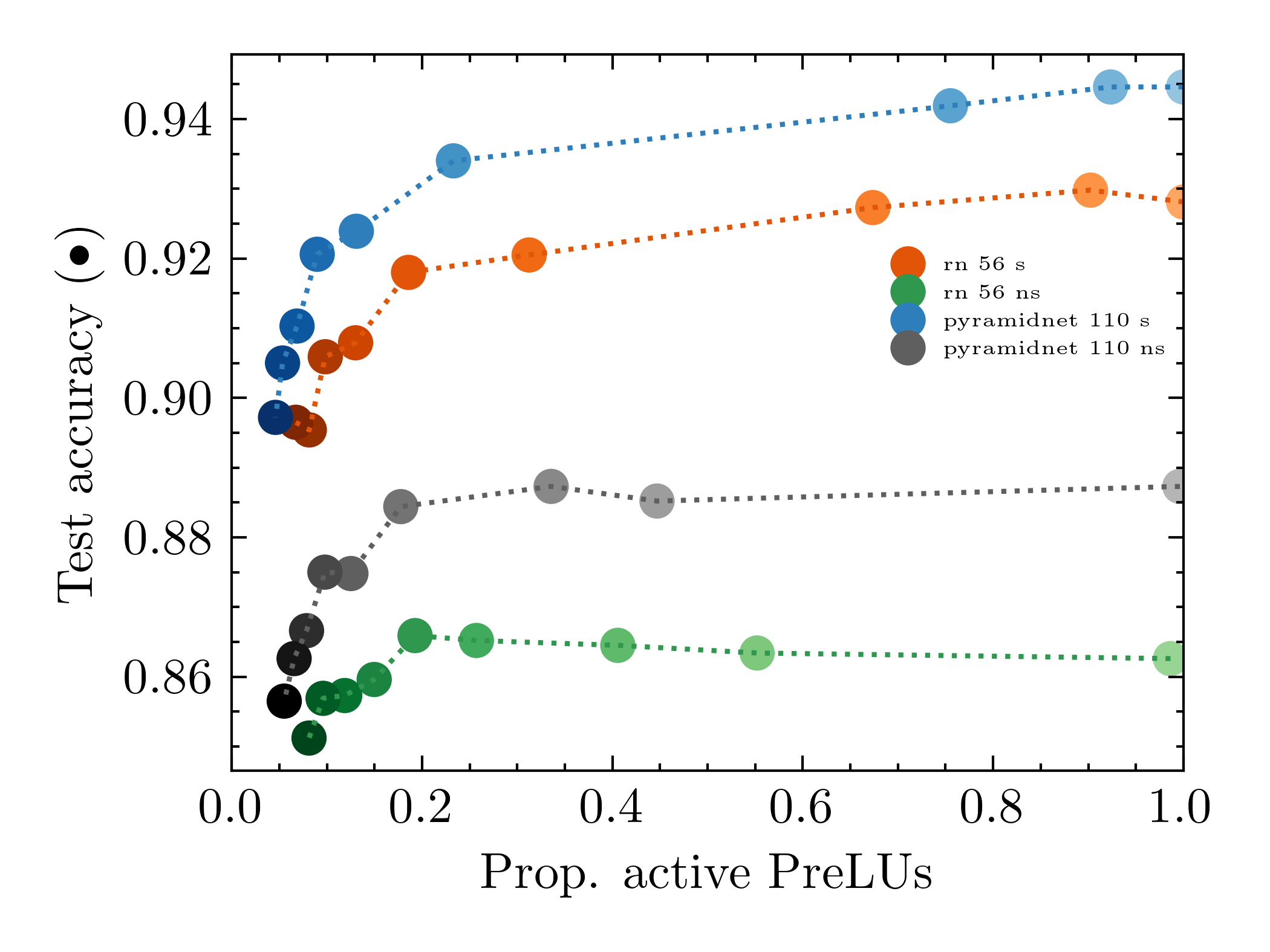}
		\caption{\textbf{Proportion of active PReLU units} and test accuracy for different network architectures with regularization weight $\omega\in [0.0005, 0.005]$.}
		\label{regtest_archcompare_active}
	\end{minipage}
	\hfill
	\begin{minipage}[t]{0.475\linewidth}\vspace{0mm}%
		\centering
		\includegraphics[width=7cm]{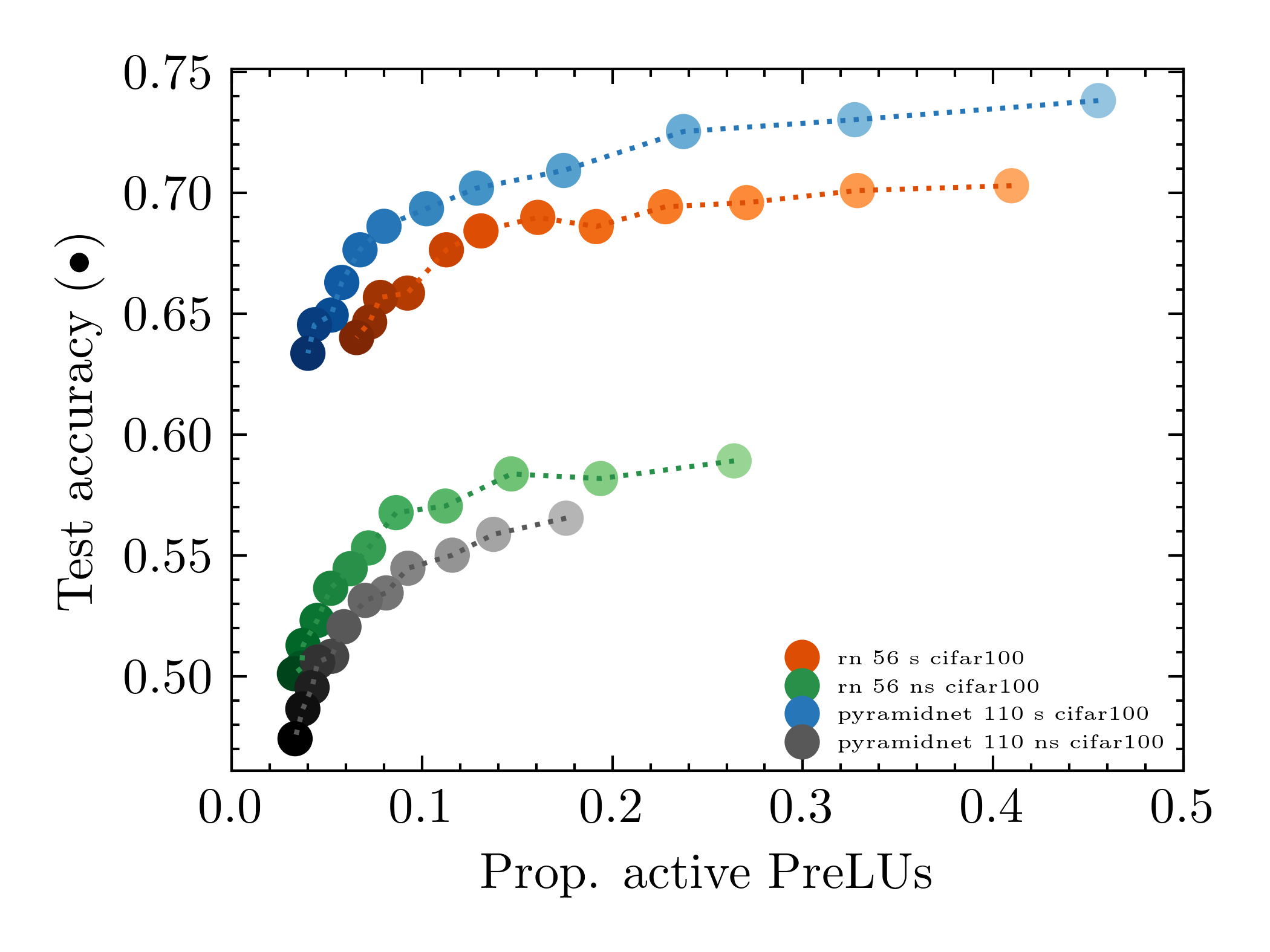}
		\caption{\textbf{Proportion of active PReLU units} and test performance (left) for different network architectures with regularization weight $\omega \in [0.0015,0.007]$ \textbf{on CIFAR-100}..}
		\label{regtest_archcompare_active_cifar100}
	\end{minipage}
\end{figure*}

In Figure \ref{regtest_archcompare_active}  (resp. \ref{regtest_archcompare_active_cifar100} for Cifar-100) show the results of Figure \ref{performance} (resp. \ref{regtest_archcompare_cifar100}) but plotting the global percentage of active PReLU units instead of NAPL on the x-axis. We see that qualitatively, we obtain a very similar result. This serves as sanity-check our our (N)APL measure, showing that the measurements made qualitatively correspond to the ones made with a much simpler measure.

\subsection{Experiments on Cifar100}

\begin{figure*}[tb]
	\begin{minipage}[t]{0.475\linewidth}\vspace{0mm}%
		\centering
		\includegraphics[width=7cm]{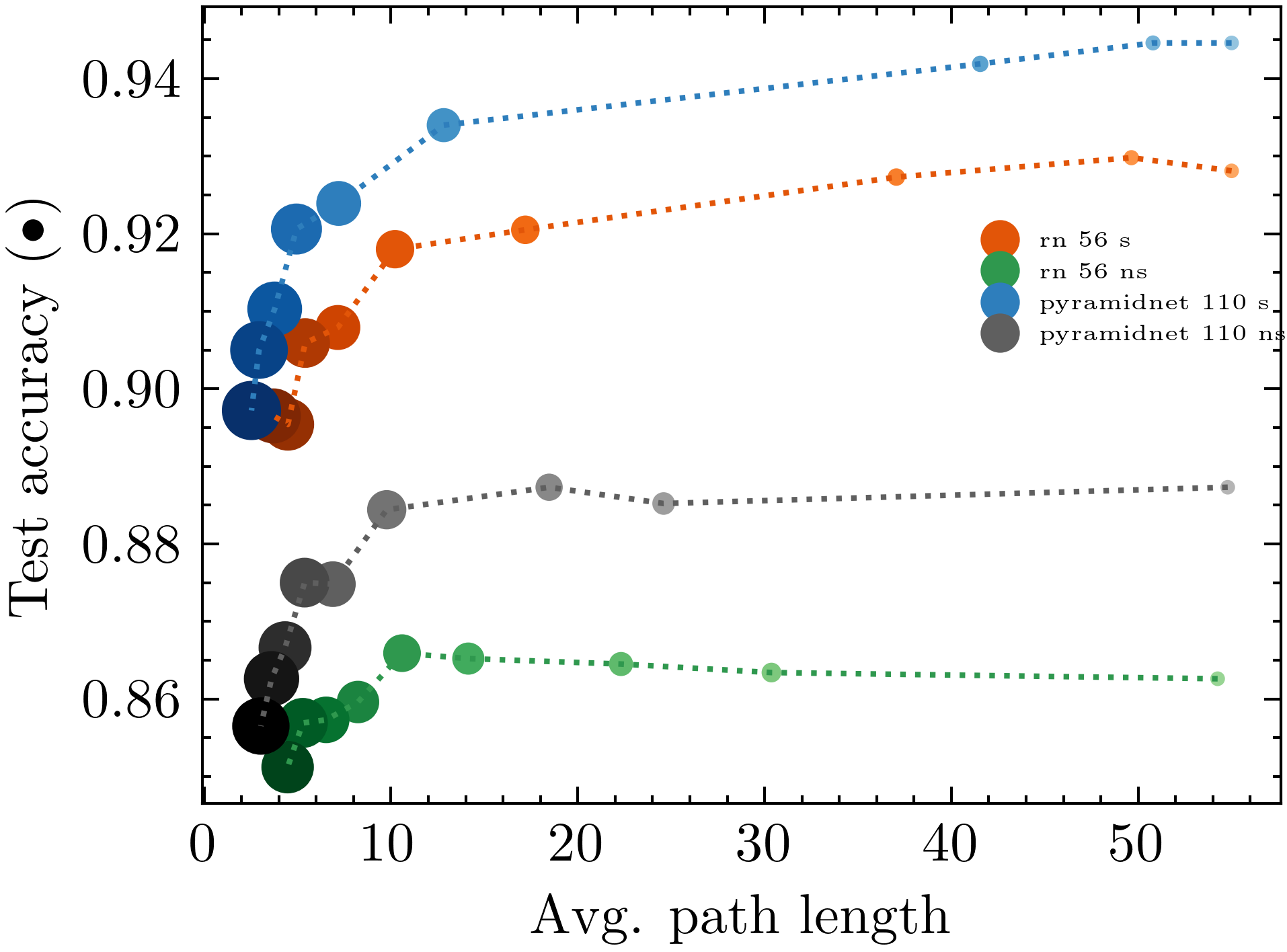}
		\caption{\textbf{Unnormalized} average path length and test accuracy for different network architectures with regularization weight $\omega\in [0.0005, 0.005]$.}
		\label{performance_apl}
	\end{minipage}
	\hfill
	\begin{minipage}[t]{0.475\linewidth}\vspace{0mm}%
		\centering
		\includegraphics[width=7cm]{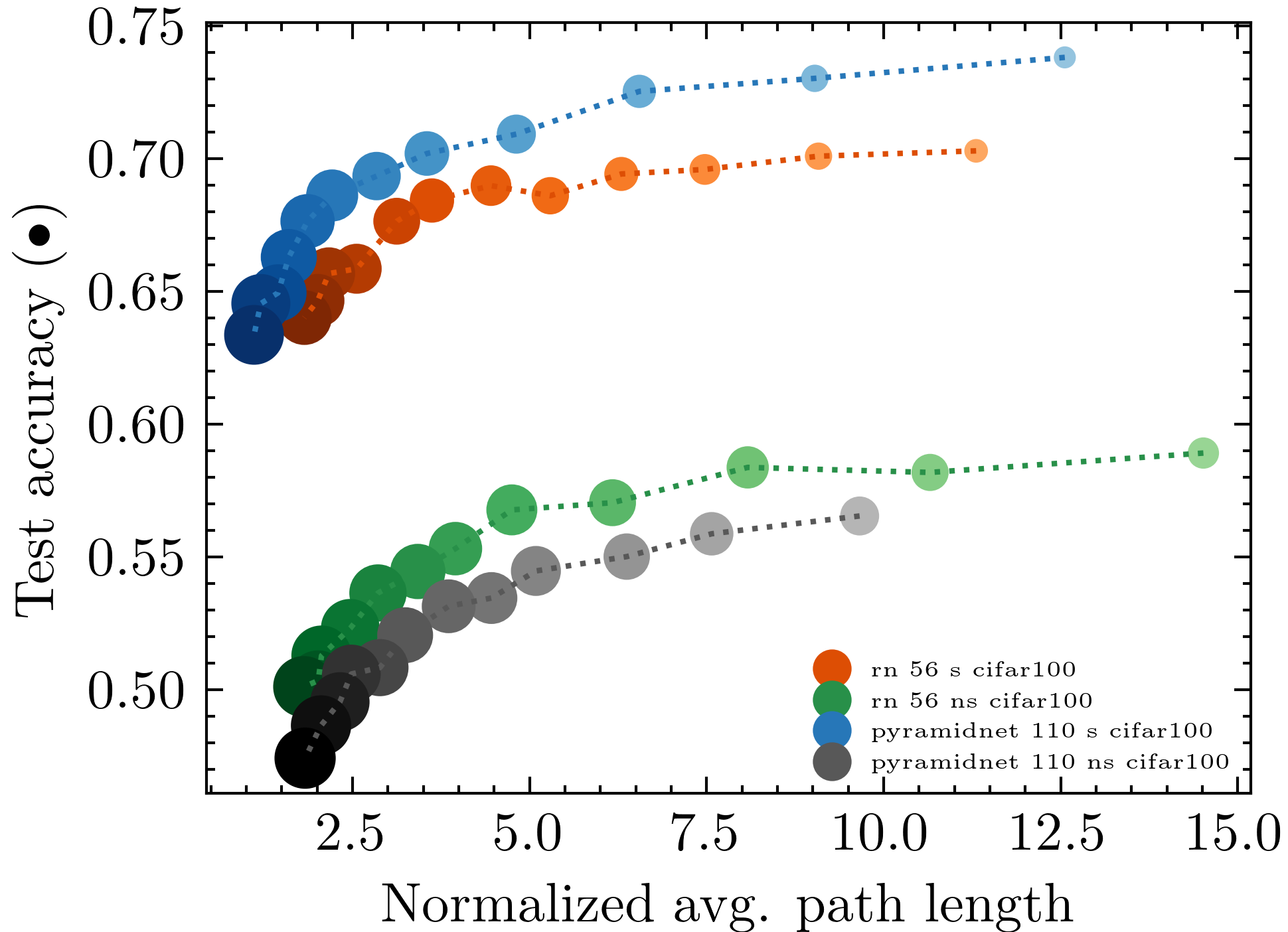}
		\caption{Normalized average path length and test performance (left) for different network architectures with regularization weight $\omega \in [0.0015,0.007]$ \textbf{on CIFAR-100}.}
		\label{regtest_archcompare_cifar100}
	\end{minipage}
\end{figure*}

\begin{figure*}[tb]
	\begin{minipage}[t]{0.475\linewidth}\vspace{0mm}%
		\centering
		\includegraphics[width=7cm]{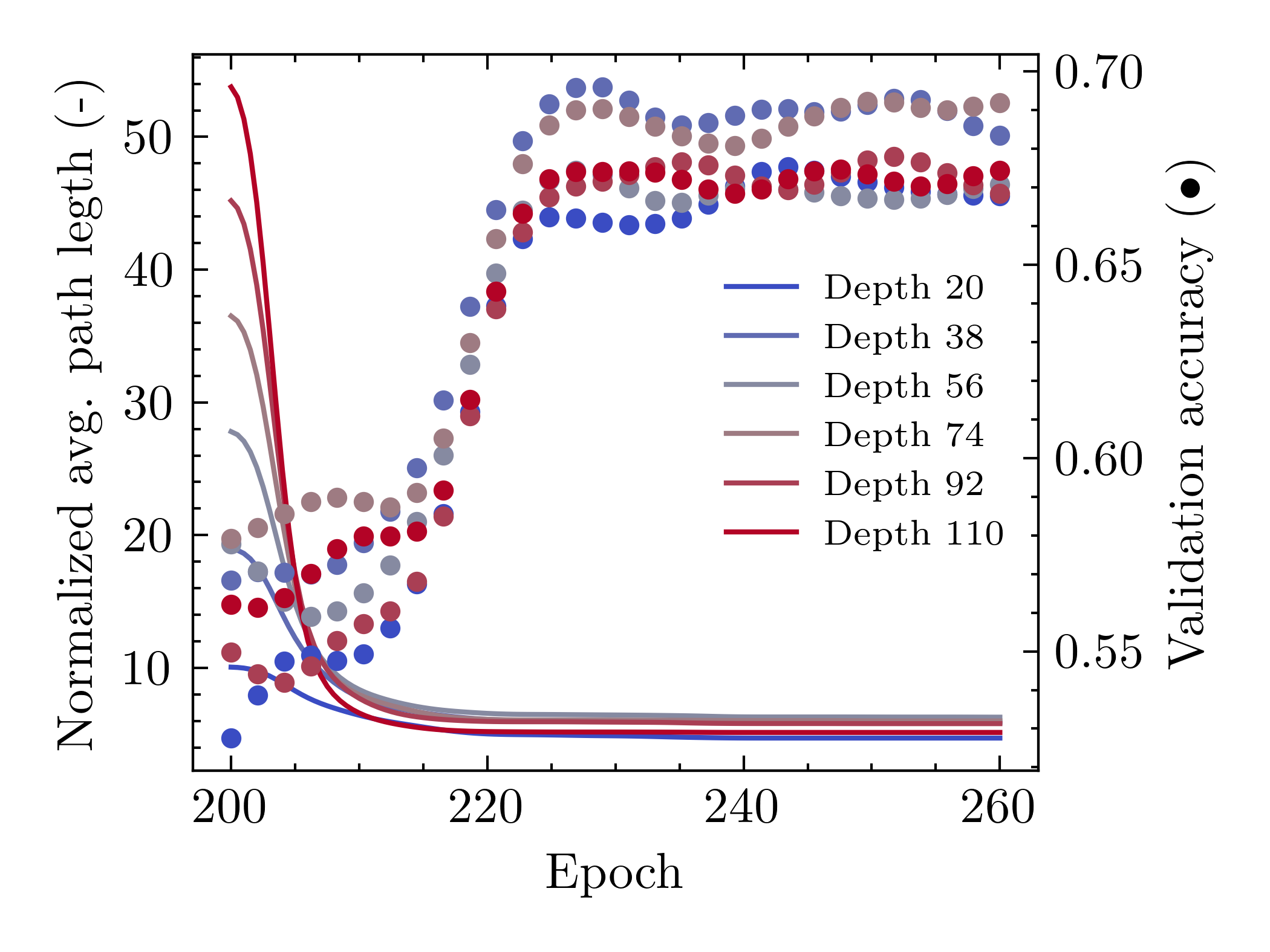}
		\caption{Validation accuracy and NAPL during linearization of ResNets of different depth for $\omega = 0.003$ \textbf{on CIFAR-100}.}
		\label{depthtest_cifar100}
	\end{minipage}
	\hfill
	\begin{minipage}[t]{0.475\linewidth}\vspace{0mm}%
		\centering
		\includegraphics[width=7cm]{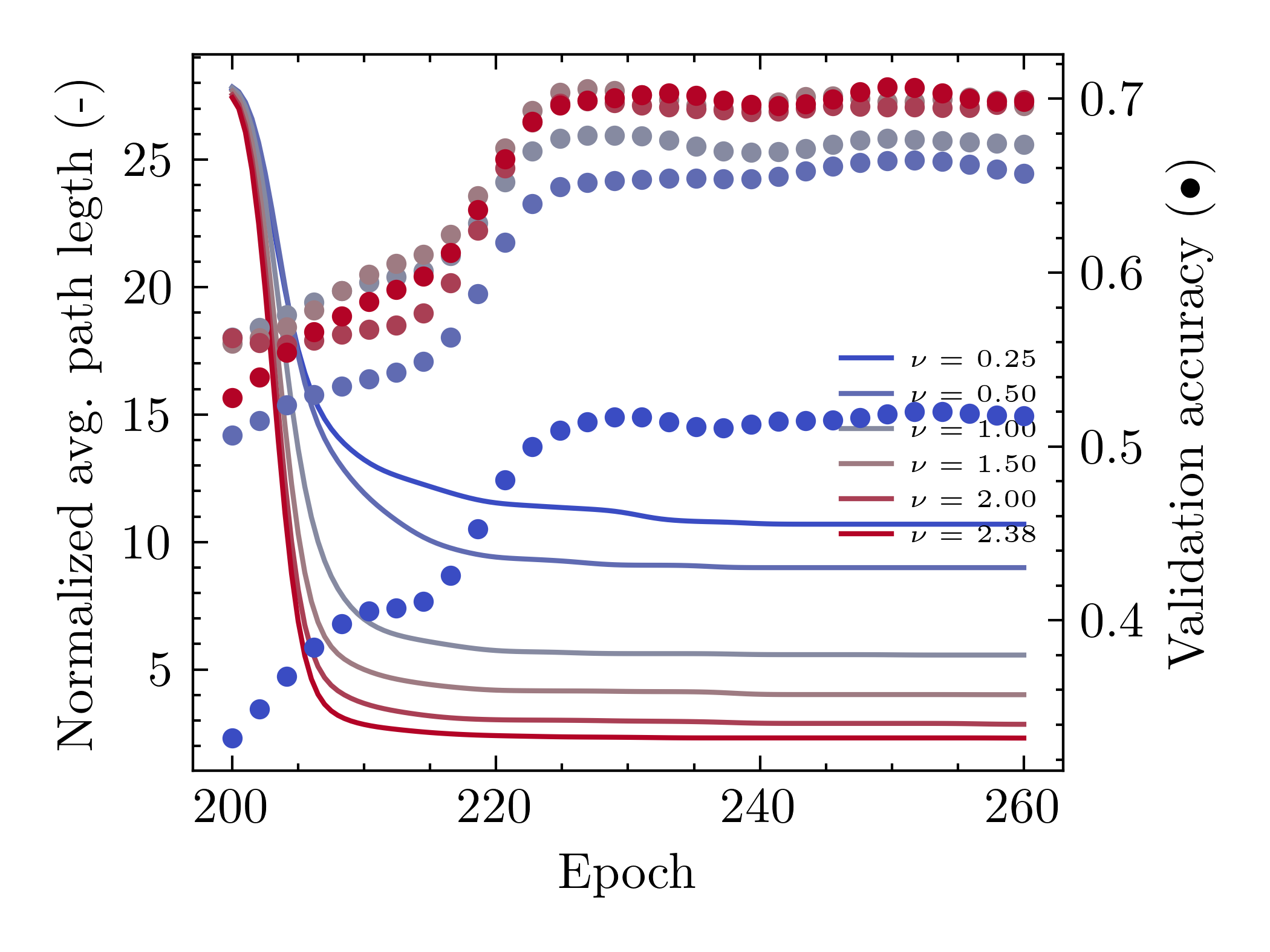}
		\caption{Validation accuracy and NAPL during linearization of ResNets56 of different width for $\omega = 0.003$ \textbf{on CIFAR-100}.}
		\label{widthtest_cifar100}
	\end{minipage}
\end{figure*}

\begin{figure*}
\begin{minipage}[t]{0.475\linewidth}
	\centering
	\includegraphics[width=0.99\linewidth]{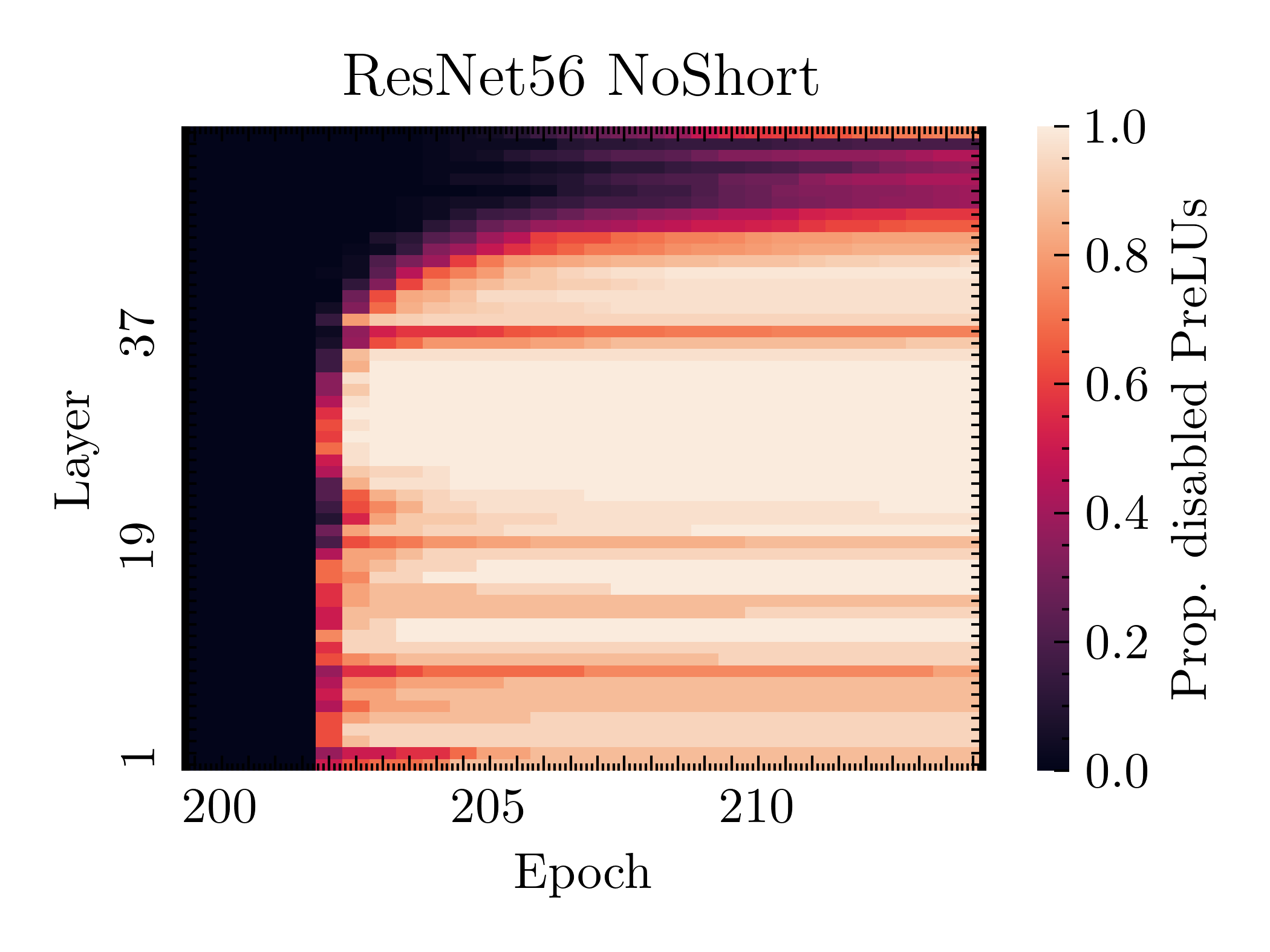}
\end{minipage}
\hfill
\begin{minipage}[t]{0.475\linewidth}
	\centering
	\includegraphics[width=0.99\linewidth]{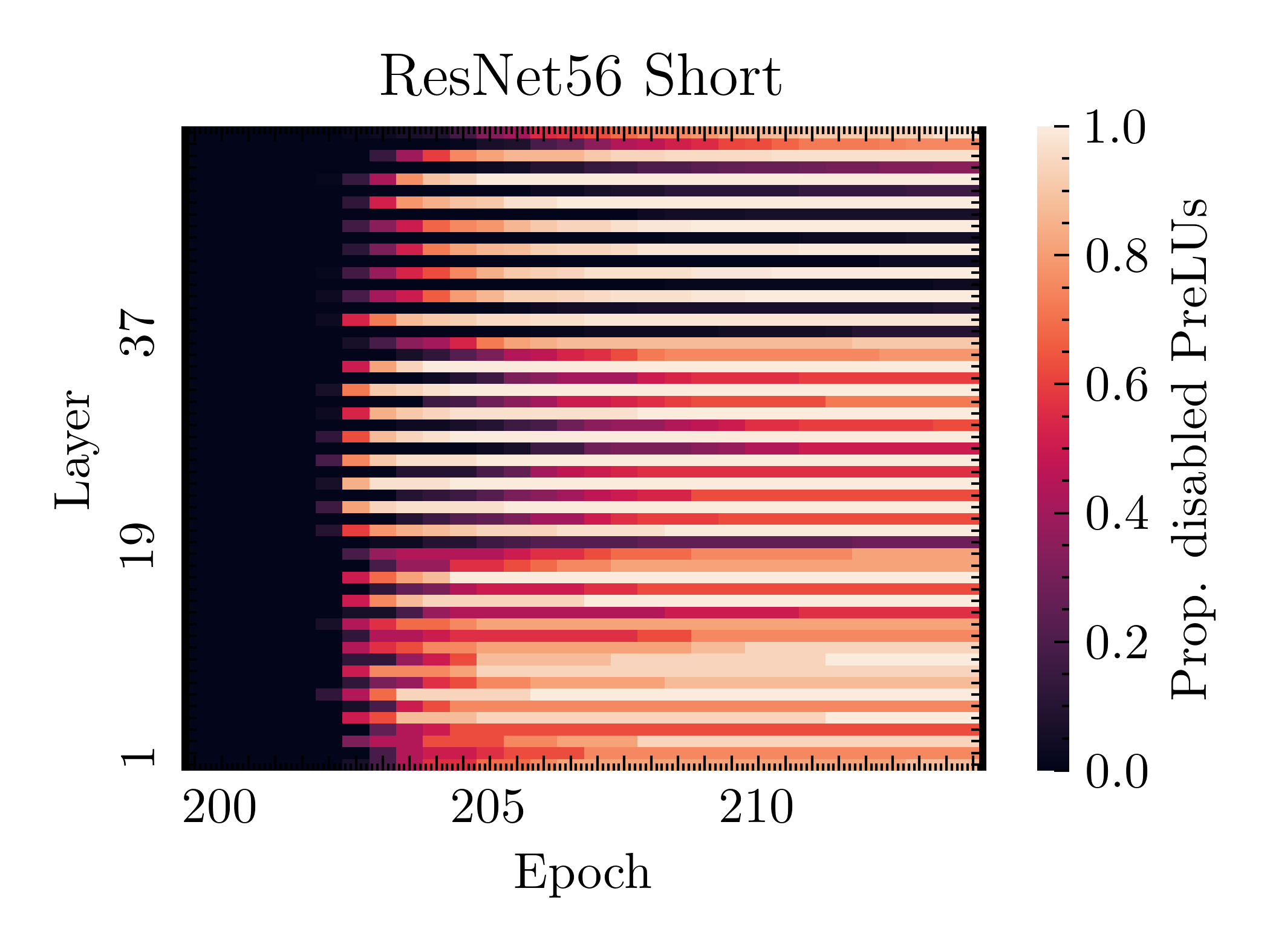}
	
\end{minipage}
\caption{Proportion of inactive PReLUs when linearizing a ResNet56 NoShort / Short with $\omega = 0.003$ \textbf{on CIFAR-100}.}
\label{pd_cifar100}
\end{figure*}

In Figure \ref{pd_cifar100}, \ref{regtest_archcompare_cifar100}, \ref{depthtest_cifar100} and \ref{widthtest_cifar100} we repeated the experiments of Figure \ref{prop_disabled}, \ref{performance}, \ref{depthtest}, \ref{widthtest} on CIFAR-100 and see qualitatively the same behaviour although all measured NAPL value are consistently higher than on CIFAR-10.

\subsection{APL of Networks with Constant Width}
\label{depthtest_ctw_section}
In Figure \ref{depthtest}, we analyzed the average path length of ResNets of different size after applying our post-training linearization procedure. In Figure \ref{depthtest} all networks seem to converge approximately to the same APL independently of network depth,  but shallower networks seem to yield a slightly higher APL. We wanted to further investigate this pattern. 

In Figure \ref{depthtest_ctw_toynet}, we repeated the same experiment with a simplistic Toy-Net having constant width and no striding after the first layer and see the pattern dissapear completely. We conclude that the observed pattern is an artifact of scaling ResBlocks of different width and striding operations in the network.

\subsection{Nonlinear Advantage on Transformer Architectures}
\new{In Figure \ref{nonlin_advantage_nlp}, we repeated the experiment of Figure \ref{nonlin_advantage} for a transformer network training on the Multi30k german-english translation task. We decided to linearize $100\%$ of all PReLU activations, as transformer architectures contain more nonlinearities than just ReLU units and we need to remove enough nonlinearity in the network in order to significantly impact performance. We see the result of the main section confirmed: networks linearized at a later stage of training outperform networks linearized earlier on.}

\section{Architecture and Training Details}
\label{appendix:sec:arch_details}
\subsection{Architecture Details}
As described in the main paper, we used a ResNet architecture with BasicBlock v2 and BasicBlockPyramid with and without residual connections for the CIFAR-10 / CINIC-10 / CIFAR-100 runs. We used shortcut option "A" (padding) for all networks except in Section \ref{section_width} where option "B" (1x1 convolutions) is needed to make the network work with different widths.  We used the default number of planes $num\_planes = (16, 32, 64)$ for each BasicBlock, except for Figure \ref{widthtest} where the number of planes is multiplied by a constant $\nu$. For Figure \ref{depthtest}, we used $num\_blocks = (i,i,i),~i\in \{3,6,9,12,15,18\}$ to scale the number of blocks in the ResNet. PyramidNet 41 resp. 110 uses  $num\_blocks = (3,4,6)$ resp. $18, 18, 18$ and $num\_planes = (32, 128)$ resp $num\_planes = (16, 100)$ with shortcut option "A".

For the ImageNet runs, we used the ResNet50 architecture from the official TorchVision repository.

\begin{figure*}[tb]
\begin{minipage}[p]{0.475\linewidth}\vspace{0mm}%
	\centering
	\includegraphics[width=7cm]{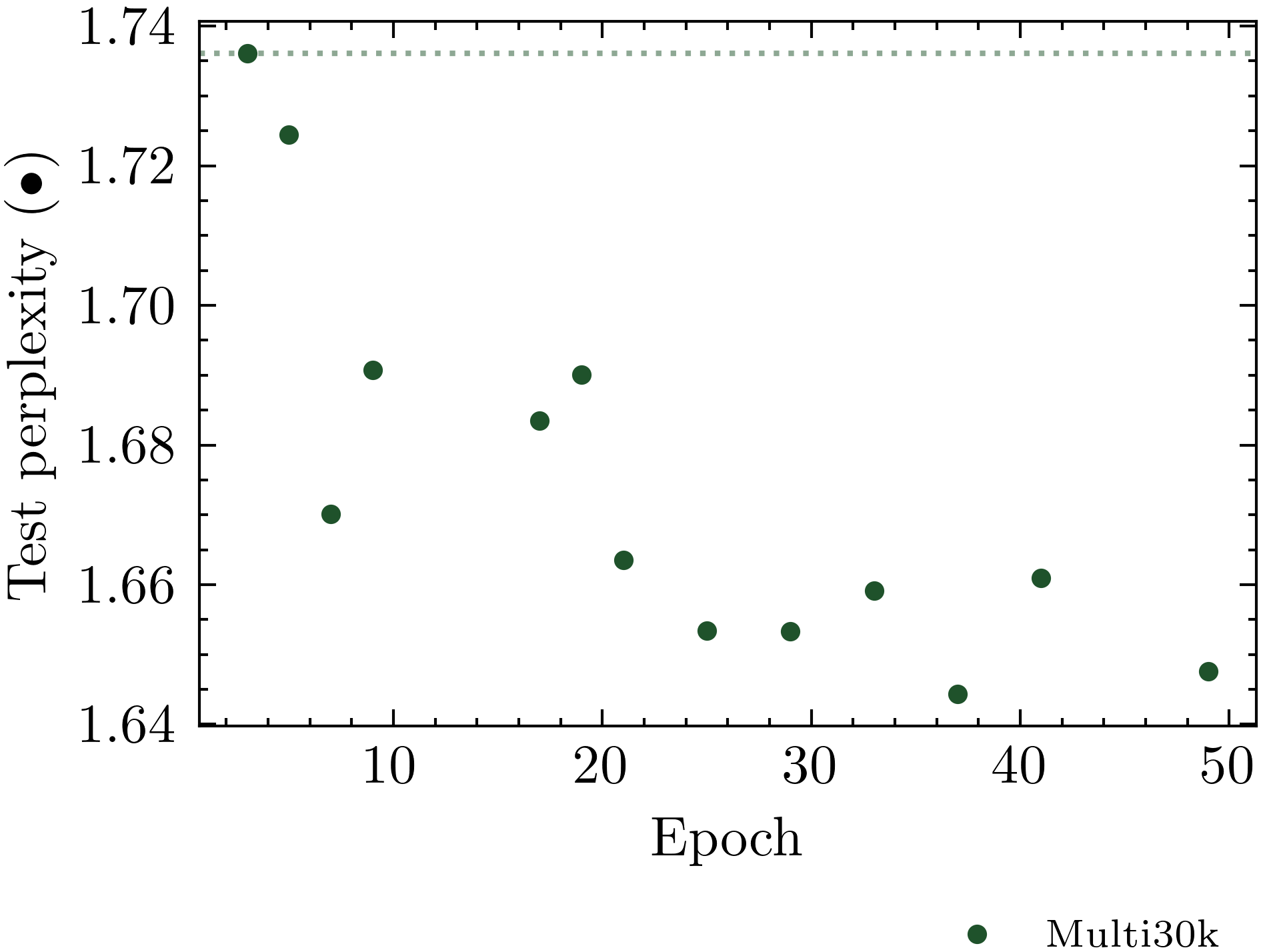}
	\caption{Test perplexity of a transformer network partially linearized at different epochs during training on a translation task. The dotted line indicates the height of the first datapoint for visual reference.}
	\label{nonlin_advantage_nlp}
\end{minipage}
\hfill
\begin{minipage}[p]{0.475\linewidth}\vspace{0mm}%
	\centering
	\includegraphics[width=7cm]{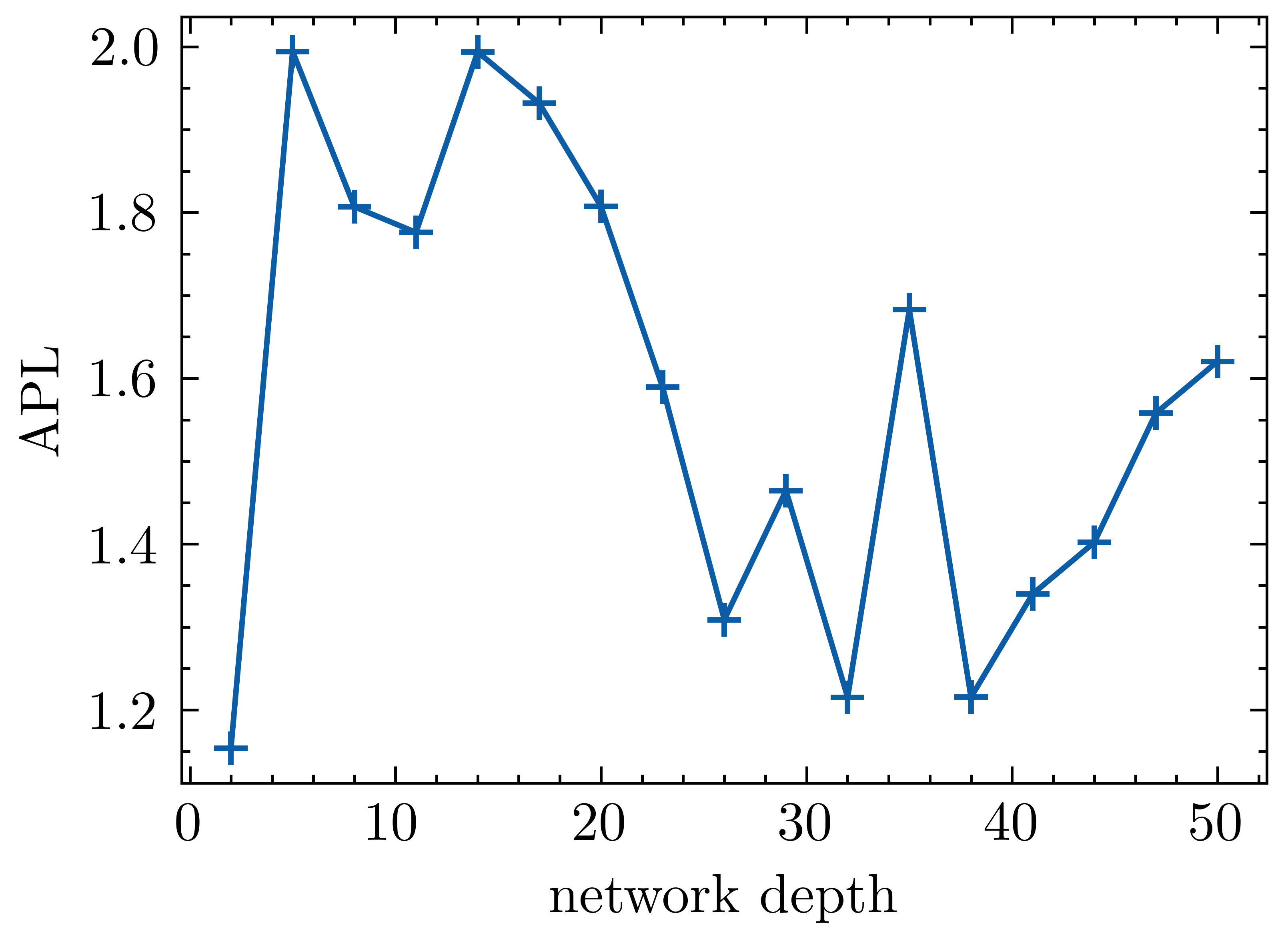}
	\caption{Average path length of Toy-Nets of different depth after post-training linearization for $\omega = 0.003$.}
	\label{depthtest_ctw_toynet}
\end{minipage}

\end{figure*}

For Figure \ref{depthtest_ctw_toynet}, we used a simple Conv-BN-ReLU ToyNet with constant width (32 Filters), no striding after the first layer, residual connections of length 1 and a final fully connected layer.

Details of the transformer architecture can be found in Figure \ref{transformer_details} (left).

\subsection{(Post-)Training Hyperparmeters \& Hardware}
The experiments in the paper were made on computers running Arch Linux, Python 3.10.5, PyTorch Version 1.11.0+cu102. The GPUs used were NVIDIA GeForce GTX 1080 Ti and NVIDIA GeForce RTX 2080 Ti.

The hyper-parameters in Figure \ref{train_table} were used to (post-) train on the CIFAR-10, CINIC-10 and CIFAR-100 and usually reach the standard test-accuracy of approximately $92.7$ for a ResNet56 on CIFAR-10. As for the ImageNet runs, we used a pre-trained model from the torchvision model-zoo. For post-training, the hyper-parameters in Figure \ref{posttrain_table} were used. For the CINIC-10 post-training, we adapted the number of epochs and the multistep scheduler milestones to approximately maintain the same number of batches since the total number of training images is different.

\textbf{Note:} The experiments of Figure \ref{retrain_table} and \ref{prop_disabled} have a shorter post-train phase of 30 epochs instead of 60 epochs (the multistep milestones are 10/20) to save compute, as these experiments do not aim for maximum accuracy.

\begin{figure}
		\begin{minipage}[p]{0.475\linewidth}\vspace{0mm}%
		\begin{tabular}{ l  l }
	\toprule\\
	Architecture& Transformer \\
	\midrule\\
	$d_{ff}$& 2048\\
	$d_{model}$&512\\
	$h$&8\\
	$N$&6\\
	$p_{dropout}$&0.1\\
	\bottomrule
		\end{tabular}
	\end{minipage}
	\hfill
	\begin{minipage}[p]{0.475\linewidth}\vspace{0mm}%
		\begin{tabular}{ l  l }
	\toprule\\
	Training& Multi30k \\
	\midrule\\
	Epochs& 200\\
	Scheduler&Warmup+Multistep ($\gamma=0.1$)\\
	Warmup steps&2900\\
	Milestones&160, 180\\
	Learning rate&0.00082\\
	Batch size&30 \\
	Gradient accumulation&10 steps \\
	Optimizer& ADAM\\
	$(\beta_1, \beta_2)$&(0.9, 0.98)\\
	Weight decay& 0\\
	Label Smoothing& 0.1\\
	\bottomrule
		\end{tabular}
	\end{minipage}
	\caption{Architecture details and training regime for the NLP task.}
	\label{transformer_details}
\end{figure}

\begin{figure*}[tb]
	\begin{minipage}[t]{0.475\linewidth}\vspace{0mm}%
		\begin{tabular}{ l  l }
			\toprule\\
			Training& CIFAR-10 / CINIC-10 / CIFAR-100\\
			\midrule\\
			Epochs& 200\\
			Scheduler&Multistep ($\gamma=0.1$)\\
			Milestones&100, 150\\
			Learning rate&0.1\\
			Batch size&256\\
			Optimizer& SGD + Momentum\\
			Momentum&0.9\\
			Weight decay& 0.0001\\
			Augmentation& Random Flip\\
			\bottomrule
		\end{tabular}
	\end{minipage}
	\hfill
	\begin{minipage}[t]{0.475\linewidth}\vspace{0mm}%
		\begin{tabular}{ l  l }
			\toprule\\
			Training& TinyImagenet\\
			\midrule\\
			Epochs& 80\\
			Scheduler&Multistep ($\gamma=0.1$)\\
			Milestones&70, 75\\
			Learning rate&0.1\\
			Batch size&128\\
			Optimizer& SGD + Momentum\\
			Momentum&0.9\\
			Weight decay& 0.0001\\
			Augmentation& Random Flip\\
			\bottomrule
		\end{tabular}
	\end{minipage}
	\caption{Details of the training regime for CV tasks.}
	\label{train_table}
\end{figure*}

\begin{figure*}[]
	\begin{minipage}[t]{0.475\linewidth}\vspace{0mm}%
		\begin{tabular}{ l  l }
			\toprule\\
			Post-Training& CIFAR-10 / CINIC-10 / CIFAR-100\\
			\midrule\\
			Epochs& 60 (34 for CINIC-10)\\
			Scheduler&Multistep ($\gamma=0.1$)\\
			Milestones&20, 40 (10, 22 for CINIC-10)\\
			Learning rate&0.1\\
			Batch size&256\\
			Optimizer& SGD + Momentum\\
			Momentum&0.9\\
			Weight decay& 0.0001\\
			Augmentation& Random Flip\\
			\bottomrule
		\end{tabular}
	\end{minipage}
	\hfill
	\begin{minipage}[t]{0.475\linewidth}\vspace{0mm}%
		\begin{tabular}{ l  l }
			\toprule\\
			Post-Training& Imagenet\\
			\midrule\\
			Epochs& 6\\
			Scheduler&Multistep ($\gamma=0.1$)\\
			Milestones& 2, 4\\
			Learning rate&0.01\\
			Batch size&40\\
			Optimizer& SGD + Momentum\\
			Momentum&0.9\\
			Weight decay& 0.0001\\
			Augmentation& Center Crop (224 px.)\\
			\bottomrule
		\end{tabular}
	\end{minipage}
	\caption{Details of the post-training regime.}
	\label{posttrain_table}
\end{figure*}

\end{document}